\PassOptionsToPackage{table}{xcolor}

\documentclass[journal]{IEEEtran}
\usepackage[most]{tcolorbox}
\IEEEoverridecommandlockouts

\usepackage{tikz}
\usepackage{textcomp}

\usepackage{lipsum}

\newcommand\copyrighttext{%
  \footnotesize \textcopyright 2025 IEEE. Personal use of this material is permitted.
  Permission from IEEE must be obtained for all other uses, in any current or future
  media, including reprinting/republishing this material for advertising or promotional
  purposes, creating new collective works, for resale or redistribution to servers or
  lists, or reuse of any copyrighted component of this work in other works.
  }
\newcommand\copyrightnotice{%
\begin{tikzpicture}[remember picture,overlay]
\node[anchor=south,yshift=5pt] at (current page.south) {\fbox{\parbox{\dimexpr\textwidth-\fboxsep-\fboxrule\relax}{\copyrighttext}}};
\end{tikzpicture}%
}
\usepackage{caption}
\usepackage{subcaption}
\usepackage{cite}
\usepackage{hyperref}
\hypersetup{
 linktocpage=true,
 pdfborderstyle={/S/S/W 1},
 hyperindex=true,
 bookmarks=true,
 bookmarksopen=true,
 bookmarksnumbered=true,
}

\usepackage{graphics} % for pdf, bitmapped graphics files
\usepackage{epsfig} % for postscript graphics files
\usepackage{mathptmx} % assumes new font selection scheme installed
\usepackage{times} % assumes new font selection scheme installed
\usepackage{amsmath} % assumes amsmath package installed
\usepackage{amssymb}  % assumes amsmath package installed
\usepackage{makecell} % For creating multiline cells
\usepackage{threeparttable}
\usepackage{adjustbox}

\usepackage{booktabs}
\usepackage{multirow}
\usepackage{float}
\usepackage{booktabs}
\usepackage{multirow}

\definecolor{darkgreen}{rgb}{0.0, 0.5, 0.0}

\ifCLASSINFOpdf
\else
\fi

\newcommand{\greyrule}{\arrayrulecolor{black!30}\midrule\arrayrulecolor{black}}

\begin{document}

\title{DriveSOTIF: Advancing SOTIF Through Multimodal  Large Language Models}
% DriveSOTIF:  Muti-modal  Large Language Models in SOTIF Research Through Self-Alignment 

\author{ Shucheng~Huang$^{1,2,3}$,~
        Freda~Shi$^{2,3}$,~
        Chen~Sun$^{4*}$,~
        Jiaming~Zhong$^{1}$,~
        Minghao Ning$^{1}$,~
        Yufeng Yang$^{1}$,~
        Yukun Lu$^{5}$,~
                Hong Wang$^{6}$,~\IEEEmembership{Senior Member,~IEEE,}
        and Amir~Khajepour$^{1}$,~\IEEEmembership{Senior Member,~IEEE}

% \thanks{Copyright (c) 2025 IEEE. Personal use of this material is permitted. However, permission to use this material for any other purposes must be obtained from the IEEE by sending a request to pubs-permissions@ieee.org.}

\thanks{Manuscript received 2 February, 2025; revised 27 August, 2025; accepted 7 September, 2025.}

\thanks{$^{1}$S. Huang, J. Zhong, M. Ning, Y. Yang, and A. Khajepour are with MVSLab, Department of Mechanical and Mechatronics Engineering, University of Waterloo, 200 University Ave West, Waterloo ON, N2L3G1 Canada.}
% % Hong Wang are from 
\thanks{$^{2,3}$S. Huang, and F. Shi are with CompLING Lab, David R. Cheriton School of Computer Science, University of Waterloo, 200 University Ave West, Waterloo ON, N2L3G1 Canada and Vector Institute, Toronto, Canada}

\thanks{$^{4}$C. Sun is with the Department of Data and Systems Engineering, University of Hong Kong, Pok Fu Lam, Hong Kong, China (e-mail: c87sun@hku.hk)}
\thanks{$^{5}$Y. Lu is with the Department of Mechanical Engineering, University of New Brunswick, Fredericton, NB E3B 5A3, Canada (e-mail: yukun.lu@unb.ca)}
\thanks{$^{6}$H. Wang is with School of Vehicle and Mobility, Tsinghua University, Beijing, China, 100084. (e-mail: hong\_wang@tsinghua.edu.cn)
}
\thanks{\textsuperscript{*} Corresponding author: Chen Sun}

}

\maketitle
\copyrightnotice

\begin{abstract}
Human drivers possess spatial and causal intelligence, enabling them to perceive driving scenarios, anticipate hazards, and react to dynamic environments. In contrast, autonomous vehicles lack these abilities, making it challenging to manage perception-related Safety of the Intended Functionality (SOTIF) risks, especially under complex or unpredictable driving conditions.
To address this gap, we propose fine-tuning multimodal large language models (MLLMs) on a customized dataset specifically designed to capture perception-related SOTIF scenarios. Benchmarking results show that fine-tuned MLLMs achieve an 11.8\% improvement in close-ended VQA accuracy and a 12.0\% increase in open-ended VQA scores compared to baseline models, while maintaining real-time performance with a 0.59-second average inference time per image.
We validate our approach through real-world case studies in Canada and China, where fine-tuned models correctly identify safety risks that challenge even experienced human drivers. This work represents the first application of domain-specific MLLM fine-tuning for the SOTIF domain in autonomous driving. The dataset and related resources are available at \url{github.com/s95huang/DriveSOTIF.git}
\end{abstract}


\IEEEpeerreviewmaketitle

\section{Introduction}

\IEEEPARstart{I}{n} autonomous driving (AD), safety is commonly classified into functional safety and Safety of the Intended Functionality (SOTIF). Functional safety concerns failures in hardware or software that result in unsafe operation. In contrast, SOTIF addresses hazards that occur not due to malfunctions, but when the system operates as intended yet produces unsafe outcomes because of external factors or inherent limitations \cite{wang2023holistic}.

Perception systems in autonomous vehicles (AVs), which are tasked with detecting, classifying, and predicting based on environmental stimuli, are particularly vulnerable to SOTIF-related challenges. Issues such as sensor limitations in adverse weather, difficulty in recognizing unusual road users, or interpreting rare and complex scenarios can lead to unsafe decisions \cite{wang2024survey}. The unpredictable nature of real-world driving, such as sudden obstacles or erratic road users, can easily overwhelm current algorithms \cite{peng2023sotif}.

Detecting and addressing SOTIF risks are therefore essential for the safe operation of AVs. Existing methods primarily adopt rule-based approaches to quantify perception-related risks by measuring entropy levels and classifying them into specific categories \cite{peng2023sotif,peng2023pesotif}. Learning-based methods, on the
other hand, suffer from black-box behavior, poor generalization to edge cases, and vulnerability to adversarial conditions. In contrast, human drivers can naturally assess the risk and uncertainty with certain objects when driving.  This capability of assessing SOTIF-related risk and selecting the best action based on personal experience,  knowledge, common sense remains absent from existing literature.

Recent advances in multimodal large language models (MLLMs) offer new approaches for addressing these cognitive gaps. MLLMs provide three unique capabilities \cite{wu2022multi, naik2023context}:
\begin{itemize}
    \item \textbf{Open-world generalization:} leveraging pre-trained knowledge to interpret unseen or novel scenarios.
\item \textbf{Causal reasoning:} tracing cause–effect relationships to understand how individual factors, or their combinations, contribute to specific outcomes.
    \item \textbf{Contextual understanding:} interpreting meaning within specific contexts, such as social interactions and human behavioral patterns.
\end{itemize}
For example, an MLLM may infer that sun glare obscures lane markings or that rain, road construction, and nighttime driving jointly increase safety risks. Importantly, MLLMs can assess such scenarios and provide human-readable explanations of the hazards involved, along with recommended actions.

\begin{table*}[ht]
% \small
\centering
\caption{Perception-Related SOTIF Datasets}

\begin{tabular}{lcccccp{8cm}}
\toprule
\textbf{Dataset} & \textbf{Year} & \textbf{Size} & \textbf{Image} & \textbf{LiDAR} & \textbf{Text} & \textbf{Description} \\
\midrule
\textbf{PeSOTIF\cite{peng2023pesotif}} & 2022 & 1126 frames & \checkmark &  &  & 2D object detection dataset for long-tail traffic scenarios with diverse weather conditions and uncommon object appearances. \\
 \greyrule
\textbf{CODA\cite{li2022coda}} & 2022 & 1500 scenes & \checkmark &  &  & 2D object detection dataset for road corner case  with 34 object categories. \\
\greyrule
\textbf{SOTIF-PCOD \cite{vehits24}} & 2024 & 547 frames &  & \checkmark &  & Sythetic LiDAR point cloud dataset generated using CARLA simulator. Contain 21  weather conditions and times of day for SOTIF-related use cases. \\
 \greyrule
\textbf{CODA-LM \cite{li2024automated}} & 2024 & 1500 scenes  & \checkmark &  & \checkmark & VQA dataset generated by extending CODA with structured textual descriptions for general perception, regional perception, and driving suggestion. \\
\bottomrule
\end{tabular} 
\label{tab:sotif_datasets}
\end{table*}

Motivated by human drivers' cognitive abilities, we propose a novel framework that leverages MLLMs to address perception-related SOTIF risks. Our approach fine-tunes pre-trained MLLMs on \textit{DriveSOTIF}, a domain-specific dataset carefully designed to capture safety-critical driving scenarios. Through supervised fine-tuning (SFT), we enhance these models with specialized knowledge that allows them to detect, reason about, and effectively respond to safety-critical driving situations. This approach directly improves situational awareness, risk assessment, and decision-making for safer and more reliable autonomous vehicles.
The key contributions of this study are summarized as follows:

\begin{enumerate}
\item \textbf{Novel Integration of MLLMs into SOTIF Research:} We introduce a comprehensive approach that integrates pre-trained MLLMs with supervised fine-tuning (SFT) on a tailored, domain-specific dataset. Our method enhances the capabilities of these models to accurately detect, reason about, and effectively mitigate perception-related SOTIF risks.

        \item \textbf{DriveSOTIF Dataset:} We introduce DriveSOTIF, the first visual question answering and image captioning dataset for perception-related SOTIF challenges. Our novel multi-agent generation pipeline reduces annotation costs while maintaining high quality, providing a scalable approach for developing MLLM datasets for autonomous driving.

    \item \textbf{Deployment and Validation:} Fine-tuned MLLMs were  deployed and the effects of quantization  and model size on real-time performance and compute requirements have been studied. Validation was conducted using data from the DriveSOTIF dataset, as well as real-world scenarios in Canada and China.

\end{enumerate}

% The remainder of this paper is organized as follows: Section II reviews related work on perception-related SOTIF risks and MLLM applications in autonomous driving. Section III describes the dataset generation pipeline and the DriveSOTIF dataset. Section IV details the MLLM fine-tuning process and benchmarking results. Section V presents real-time performance evaluation and real-world case studies demonstrating the effectiveness of the proposed approach in handling complex SOTIF scenarios. Section VI discusses deployment considerations and future work, while Section VII concludes the paper.

% \section{Literature Review }

\section{Literature Review}

\subsection{Perception-Related SOTIF}

Perception SOTIF risks arise when an autonomous vehicle's perception system functions correctly at the hardware and software levels but still fails to accurately interpret the driving environment due to external factors \cite{peng2023sotif}. Since perception is fundamental for downstream decision-making and planning, any misinterpretation can propagate through the system, leading to unintended emergency braking, unsafe lane changes, or even collisions.  These risks are further amplified by factors like sensor degradation due to rain, fog, snow, and glare, and uncommon object appearances \cite{peng2023pesotif}. Identifying and mitigating perception-related SOTIF risks is crucial for maintaining vehicle safety, particularly when an autonomous vehicle operates outside its predefined Operational Design Domain (ODD) \cite{jiang2024enhancing, chu2023sotif}.

Current approaches to detect and mitigate perception-related SOTIF risks can be classified into rule-based and learning-based methods. Rule-based approaches, such as Fault Tree Analysis (FTA) \cite{carlan2024sotif}, System Theoretic Process Analysis (STPA) \cite{zhang2021hazard, menekcse2024safety}, and formal verification \cite{zhao2022formal}, attempt to identify perception-triggering conditions and system failures through structured models \cite{menekcse2024safety}. However, these methods struggle with dynamic and unpredictable environments, limiting their applicability to real-world driving conditions \cite{adee2021systematic}.

In contrast, learning-based and probabilistic uncertainty quantification techniques aim to detect and mitigate perception-related SOTIF risks through data-driven models. Methods such as Bayesian  Risk Decomposition \cite{yu2025decomposition}, deep ensemble \cite{peng2023pesotif,peng2023sotif,yang2023online}, and sensor fusion techniques have been explored to quantify uncertainty in perception outputs and improve reliability \cite{carlan2024sotif}. 
Despite these advancements, learning-based methods often suffer from black-box behavior, making it challenging to interpret the underlying process, and poor generalization to edge cases, particularly in long-tail traffic scenarios.

\subsection{MLLMs for SOTIF Applications}

Multimodal large language models (MLLMs) integrate  visual feature encoders with text-based LLMs to enable cross-modal understanding and reasoning\cite{agarwal2020towards,wu2022multi, naik2023context}. By combining image and textual information from other modules, such as vehicle states, safety, decision-making, planning, and control, the system can make more informed responses and better understand and adapt to dynamic driving scenarios \cite{cui2023drivellm,chen2024driving,long2024vlm}.  These capabilities make MLLMs suitable for perception-SOTIF scenarios in unpredictable driving conditions \cite{wantiez2023scene,qian2024nuscenes,atakishiyev2023explaining,zheng2024gpt, gopalkrishnan2024multi,chen2024driving}.

\begin{figure*}[t]
    \centering
    \includegraphics[width=1\textwidth]{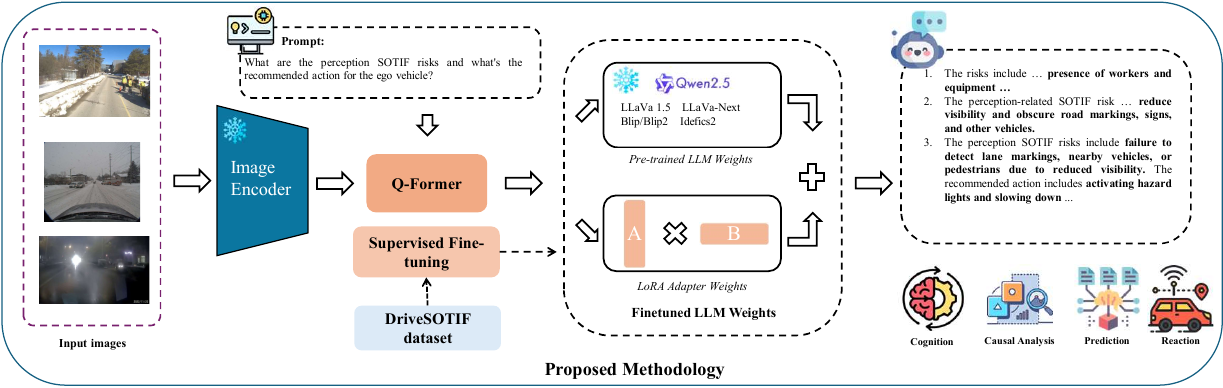}
    % \caption{Supervised fine-tuning of pre-trained multimodal LLMs on a domain-specific dataset for enhanced perception SOTIF cognition and reaction.}
    \caption{Overview of the proposed methodology for supervised fine-tuning of pre-trained MLLMs on DriveSOTIF dataset.} 
    \label{fig:methodology}
\end{figure*}

% Current SOTIF-related ML
Despite this potential, SOTIF-specific MLLM development remains limited by the datasets currently available.  As shown in Table~\ref{tab:sotif_datasets},  existing SOTIF datasets primarily focus on object detection tasks. In detail, PeSOTIF \cite{peng2023pesotif} and CODA \cite{li2022coda} are designed for 2D detection in rare and long-tail scenarios, while SOTIF-PCOD \cite{vehits24} contains synthetic LiDAR point clouds. CODA-LM \cite{li2024automated} includes text annotations for perception and driving suggestions but does not address SOTIF risk directly.

Without SOTIF-specific MLLM datasets, researchers must use general-purpose datasets with limited relevance to SOTIF challenges. Datasets like DRAMA \cite{malla2023drama} and NuScenes-QA \cite{qian2024nuscenes} are large and well-annotated but are geographically constrained, making it difficult to generalize to new scenarios. Other datasets \cite{qian2024nuscenes,wu2023referring} often use template-based question-answer generation, resulting in artificial interactions that don't reflect natural human queries. As a result, models trained/fine-tuned on these datasets struggle with open-ended reasoning and complex situational judgment.

\section{DriveSOTIF}
\label{sec:sotif_dataset}

Based on the literature review, it is clear that existing datasets fall short of supporting SOTIF-focused MLLM development. Most current resources are either centered on object detection or limited by geographic scope and template-driven annotations, which prevents them from capturing the diversity and unpredictability of real-world safety scenarios. To address this gap, we introduce DriveSOTIF, a domain-specific dataset designed to capture perception-related SOTIF scenarios for autonomous driving. As illustrated in Figure~\ref{fig:methodology}, our approach combines dataset generation with supervised fine-tuning of pre-trained MLLMs.

\subsection{Image Selection}
For this research, we utilize the first batch of images from PeSOTIF \cite{peng2023pesotif}, an object-detection dataset created to study the perception SOTIF problem in long-tail traffic scenarios. 
Specifically, the PeSOTIF dataset contains natural and artificial scenarios of perception SOTIF challenges  caused  by environmental factors such as adverse weather conditions (e.g., rain, snow, fog), challenging lighting conditions (e.g., direct sunlight, glare from headlights), and object-related factors like unusual object appearances or unexpected object behaviors. 

Today's passenger vehicles are typically equipped with multiple cameras, which requires VQA models to fuse information from different viewpoints for BEV-based perception SOTIF tasks.  For a dataset newly introduced to SOTIF, we begin with simple image-question pairs. This approach allows for comparisons with existing VQA datasets and algorithms and encourages further research in this area.

\begin{table}[t]
% \small
\centering
\caption{DriveSOTIF Dataset Answer and Question Types}
\label{tab:answer_question_type}
\begin{tabular}{l p{6.5cm} }
\toprule
\textbf{Type} & \textbf{Description} \\ \midrule
\multicolumn{2}{l}{\textbf{Answer Types}} \\ \midrule
Open-ended & No fixed question structure \\
Closed-ended & Yes/No and other limited choices \\ 
\midrule
\multicolumn{2}{l}{\textbf{Closed-ended Question Types}} \\ \midrule
Uncertainty & Uncertainty level with a certain object (Low, Med, High) \\
Existence & Does an object of class (class) exist? \\
Type & What is the type of (object) at the bottom left? \\
Counting & How many (objects) are there? \\
Key object & Is this (object) a key object? \\
\bottomrule
\end{tabular}
\end{table}

\begin{figure*}[th!]
    \centering
    \includegraphics[width=\textwidth]{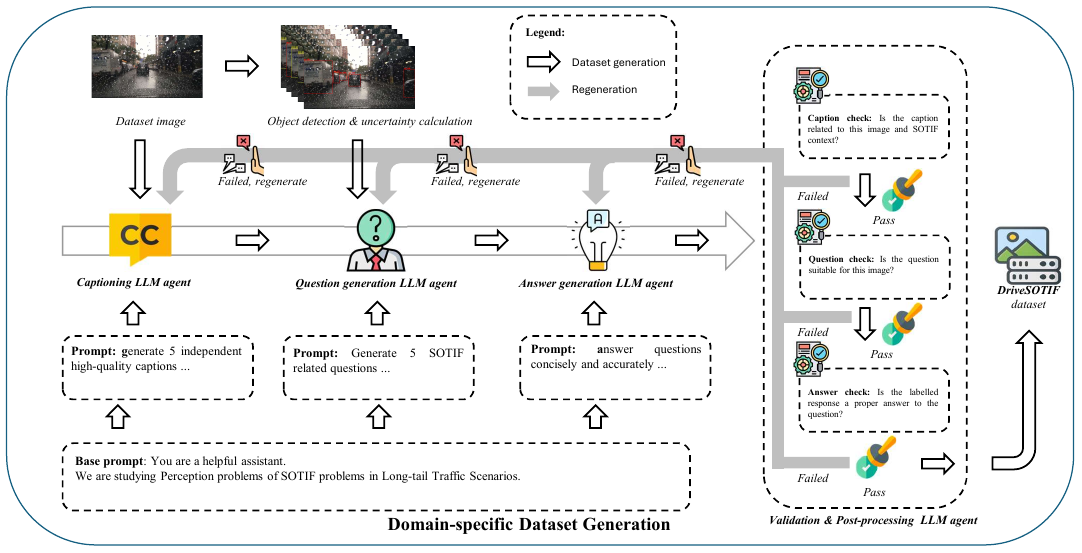}
    \caption{Proposed dataset generation pipeline through collaborative LLM Agents.}
    \label{fig:dataset_generation}
\end{figure*}

For the VQA subtask, we designed our dataset such that it supports close-ended and open-ended  Q\&A, as shown in Table~\ref{tab:answer_question_type}. Closed-ended questions are divided into five categories: uncertainty, existence, type, counting, and key object. These questions typically use templates or examples for generation, providing straightforward answers with limited choices such as yes/no or other specific options.

Open-ended Q\&A targets higher-level reasoning, including causal inference and scenario explanation. These questions do not follow a fixed structure and can have multiple correct answers, offering more flexibility and depth in responses. These questions are focused on SOTIF risk identification, explanation, and providing recommended actions.

% To ensure a balanced representation of question types, we include five questions per image, with 2–3 closed-ended and 2–3 open-ended. This ensures 
To maintain balance across question types, we assign five questions to each image, with 2–3 closed-ended and 2–3 open-ended. This design ensures balanced question diversity for comprehensive evaluation.

% \subsection{Dataset Construction}

\subsection{Dataset Generation through Collaborative LLM-Agents.}

To automatically generate the VQA and image captioning datasets, we develop a multi-agent framework using vision-capable large language models (LLMs), specifically OpenAI's GPT-4v, GPT-4o, and Anthropic's Claude 3 Opus. The dataset generation pipeline, illustrated in Figure~\ref{fig:dataset_generation}, includes distinct roles for image captioning, question generation, answer generation, and validation/post-processing. All prompts used in the dataset generation and validation process are provided in S.III of the supplemental material.

\textbf{Image Captioning Agent:}
The image captioning agent aims to generate captions based on input prompts. Importantly, we alternate  between two LLM backends for question generation and captioning. This approach is adopted to minimize bias that might arise from relying on a single proprietary LLM model. Sample images and captions are provided in Table~\ref{tbl:vqa_sample}.

\begin{table*}[th]
    % \small
    \caption{Sample labeled Questions \& Answers and image captions from SOTIF-VQA dataset.}
    \begin{center}
    \begin{tabular}{m{0.12\textwidth} m{0.4\textwidth} m{0.4\textwidth}}
    \toprule
    \textbf{Task} & \textbf{Image 1} & \textbf{Image 2} \\
    \midrule
    
    % Images
   \textbf{Image}  & \includegraphics[width=0.3\textwidth]{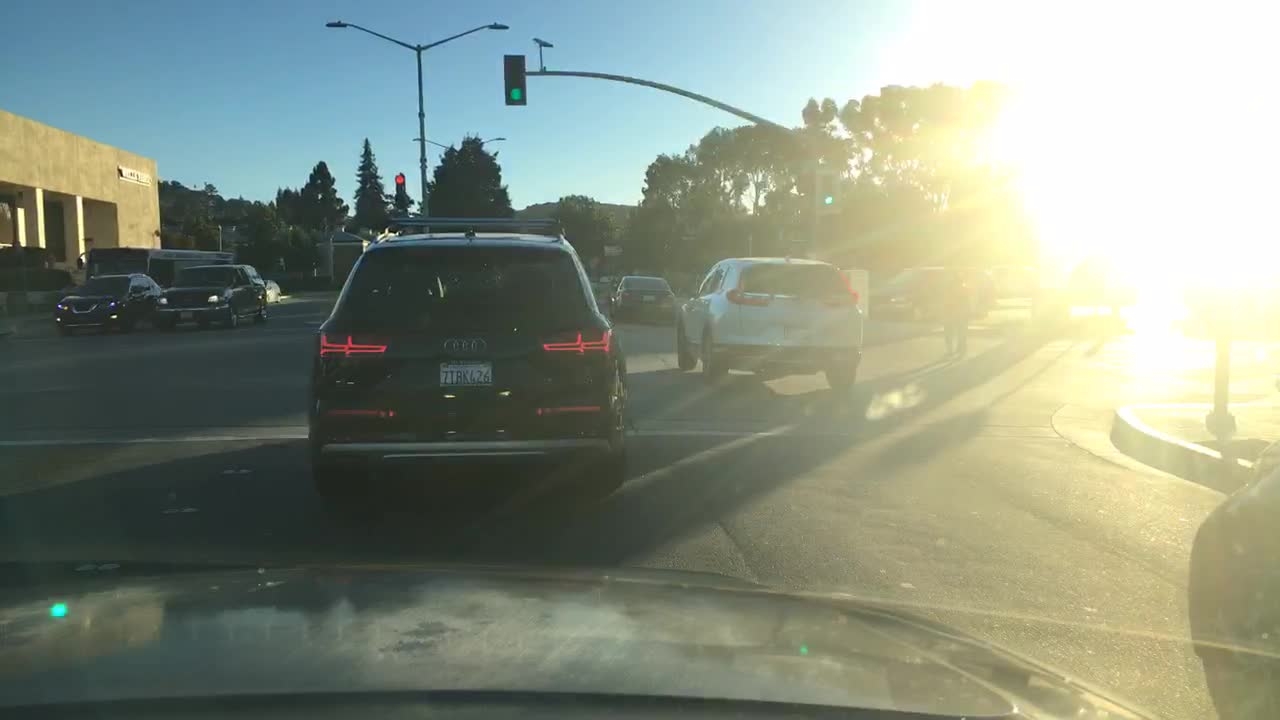} & \includegraphics[width=0.3\textwidth]{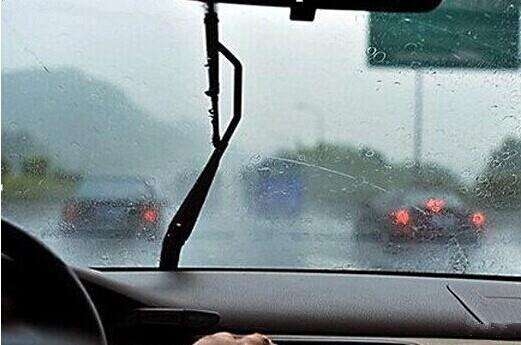} \\
    \midrule
    
    % Close-ended Questions
    \multirow{4}{*}{\textbf{Close-ended QA}} 
    & \textbf{Q1}: Is the visibility of road signs and pedestrians clear in this image? & \textbf{Q1}: How many vehicles can you discern through the windshield? \\
    & \textbf{A1}: No & \textbf{A1}: Four \\
    \cmidrule{2-3}
    & \textbf{Q2}: Does the image show a clear view of the traffic lights? & \textbf{Q2}: Is visibility reduced in this image due to weather? \\
    & \textbf{A2}: Yes & \textbf{A2}: Yes \\
    \midrule
    
    % Open-ended Questions
    \multirow{10}{*}{\textbf{Open-ended QA}} 
    & \textbf{Q}: What specific environmental factor in the image degrades the perception ability of a driver or an autonomous driving system? & \textbf{Q}: What are the perception-related SOTIF risks evident in this image due to rain? \\
    & \textbf{A}: The low sun angle causing lens flare and reduced visibility is the environmental factor degrading perception. & \textbf{A}: The rain creates a wet and reflective road surface, reduced visibility, and potential glare from surrounding vehicle lights, which can degrade the ability of an automated driving system to accurately detect lane markings, other vehicles, and pedestrians. \\
    \cmidrule{2-3}
    & \textbf{Q}: What technological improvement could help reduce the perception-related SOTIF risk due to direct sunlight? & \textbf{Q}:  What type of environment subset does this scenario belong to in the context of SOTIF problems? \\
    & \textbf{A}: Improving camera dynamic range and employing polarizing filters or advanced image processing algorithms to mitigate glare and enhance visibility could help. & \textbf{A}: This scenario belongs to the 'rain' subset within the environment category, which involves scenarios that degrade the perception ability of automated driving systems due to natural environmental factors.\\
    \midrule
    
    % Image Captioning
   \textbf{Image Captioning}
    & A driver's perspective of a busy intersection during sunset, highlighting the challenges of low sun glare in urban driving. & Driving in heavy rain, the windshield is covered in water droplets, making visibility of the traffic ahead poor. \\
    \bottomrule
    \end{tabular}
    \label{tbl:vqa_sample}
    \end{center} 
\end{table*}

\textbf{Question Generation Agent:}
The primary objective of this agent is to generate natural, human-like questions based on the input query and captions generated, along with 2D object detection and uncertainty information \footnote{Please refer to supplemental material for object detection and uncertaintyquantification. }. Specifically, the question-generating agent is instructed to generate questions with varying difficulties and types. For closed-end questions, the agent randomly selects the question type from Table~\ref{tab:answer_question_type}. For open-ended questions, the agent focuses on advanced QA like causal inference, including identifying and explaining the SOTIF risk and providing recommended action to prevent the risk, if possible. At the same time, the agent also generates metadata about the question, such as the question type, difficulty level, and expected answer type.
Sample questions generated are shown in Table~\ref{tbl:vqa_sample}.

\textbf{Answer Generation Agent:}
In the VQA dataset format, each question is answered by ten responses from ten annotators. While this approach ensures accuracy, it is both costly and time-consuming. 
 To streamline this process, the answer generation agent is designed to provide ten answers for each question, given an image. To achieve this, we employ a model aggregation method that utilizes GPT-4o and Claude3 Opus.
For each image, the image file and associated questions are fed into the LLMs.  Instead of sequentially querying a single model multiple times, the system generates five answers for each question in a single pass through each model. 
This method is more efficient and cost-effective compared to sending the questions individually. Similarly, the answer generation agent outputs metadata for each answer, including the confidence score and answer type.

\textbf{Validation and Post-processing Agent:}
As shown in Figure~\ref{fig:dataset_generation}, newly generated answers are then sent to the validation and post-processing agent, which contains three post-processing mechanisms. The first and second mechanisms are to ensure that captions and questions generated are related to the image and perception SOTIF context. The third mechanism is to ensure that answers generated are correct based on the input image and question.  If any inconsistencies are found, the agent regenerates the captions, questions or answers as needed. This validation process leverages OpenAI's GPT-4.1 as the LLM backend to ensure high-quality outputs that align with the perception SOTIF context.

\subsection{Dataset Overview}

The multi-agent system employed varying temperature settings during generation to ensure diversity across questions, answers, and captions. Following automated generation, all data entries underwent structural validation and format standardization. 

with close-ended and open-ended questions automatically categorized using metadata from the generation process. A human review of 595 samples, selected using finite population sampling\footnote{More details on the finite population sampling method can be found in S.IV of the supplemental material.} indicated a 3-4\% error rate, which is acceptable for automated pipelines.  An additional round of manual review addressed edge cases and subtle inconsistencies that automated systems might miss or generate.

The final DriveSOTIF dataset contains 1,114 images, 1,114 captions, and 5,570 question-answer pairs \footnote{Collecting SOTIF-related scenarios is challenging due to their rarity and unpredictability in real-world settings. We are actively collaborating with PeSOTIF authors to expand the dataset with new scenarios and modalities.}. The dataset adopts the COCO Caption format \cite{chen2015microsoft} for image descriptions and the VQAv2 format \cite{goyal2017making} for question-answering tasks. We partition the data using a 90/10 train-test split, as detailed in Table~\ref{tab:dataset_stats}.

% Content analysis highlights several strengths. 

Table~\ref{tbl:vqa_sample} demonstrates the dataset’s strong alignment with perception-related SOTIF challenges. Quantitatively, caption length averages 18-19 words compared to 10.5 words in COCO Caption, providing substantially richer contextual information essential for understanding complex driving scenarios. Question difficulty maintains excellent balance across easy (39.9\%), medium (40.2\%), and hard (19.9\%) categories. Similarly, question formats split evenly between close-ended (45.7\%) and open-ended (54.3\%) types, which facilitates assessment of both factual accuracy and reasoning capabilities.

Answer types span analysis (36.1\%), yes/no and multiple-choice (28.1\%), recommendations (17.8\%), counting (17.0\%), and identification (1.0\%) tasks. This diversity enables comprehensive benchmarking of models’ ability to handle both fundamental perception tasks and higher-level causal reasoning and safety recommendations specific to SOTIF.

The generation statistics demonstrate the effectiveness of the multi-agent approach and validate our quality control mechanisms. Caption generation achieved 90.1\% first-attempt success, reflecting the straightforward nature of descriptive tasks. Question generation followed with 82.7\% success, while answer generation required the most refinement, averaging 1.72 attempts per item due to the complexity of ensuring factual accuracy in analytical responses. Despite this iterative process, 71.0\% of answers passed initial validation, with 21.4\% requiring single revisions and only 7.6\% needing multiple corrections. These results demonstrate that the pipeline consistently produces high-quality, diverse, and contextually appropriate annotations for perception-related SOTIF scenarios.

\begin{table}[t]
\centering
% \small
\caption{Dataset statistics and generation information}

\begin{tabular}{@{}p{4.5cm}rr@{}}
\toprule
\multicolumn{3}{c}{\textbf{Dataset Split Statistics}} \\
\midrule
& \textbf{Train/Valid} & \textbf{Test} \\ 
\midrule
Number of questions & 5,025 & 555 \\
Number of answers & 50,250 & 5,550 \\
Number of captions & 1,005 & 111 \\
Average question length (words) & 11.9 & 11.8 \\
Average answer length (words) & 20.5 & 21.0 \\
Average caption length (words) & 19.22 & 18.22 \\ 
\midrule
\multicolumn{3}{c}{\textbf{Dataset Content Distributions}} \\
\midrule
\multicolumn{3}{l}{\textsc{Difficulty Distribution}} \\
\quad Easy questions & \multicolumn{2}{r}{2,224 \textcolor{gray}{(39.9\%)}} \\
\quad Medium questions & \multicolumn{2}{r}{2,243 \textcolor{gray}{(40.2\%)}} \\
\quad Hard questions & \multicolumn{2}{r}{1,113 \textcolor{gray}{(19.9\%)}} \\
\addlinespace[0.5em]
\multicolumn{3}{l}{\textsc{Question Type Distribution}} \\
\quad Close-ended questions & \multicolumn{2}{r}{2,549 \textcolor{gray}{(45.7\%)}} \\
\quad Open-ended questions & \multicolumn{2}{r}{3,031 \textcolor{gray}{(54.3\%)}} \\
\addlinespace[0.5em]
\multicolumn{3}{l}{\textsc{Expected Answer Type Distribution}} \\
\quad Analysis & \multicolumn{2}{r}{2,014 \textcolor{gray}{(36.1\%)}} \\
\quad Yes/No \& multiple choice & \multicolumn{2}{r}{1,568 \textcolor{gray}{(28.1\%)}} \\
\quad Recommendation & \multicolumn{2}{r}{995 \textcolor{gray}{(17.8\%)}} \\
\quad Count & \multicolumn{2}{r}{947 \textcolor{gray}{(17.0\%)}} \\
\quad Identification & \multicolumn{2}{r}{56 \textcolor{gray}{(1.0\%)}} \\
\midrule
\multicolumn{3}{c}{\textbf{Generation Information}} \\
\midrule
\multicolumn{3}{l}{\textsc{Average Generation Attempts}} \\
\quad Caption generation & \multicolumn{2}{r}{1.16 attempts/item} \\
\quad Question generation & \multicolumn{2}{r}{1.23 attempts/item} \\
\quad Answer generation & \multicolumn{2}{r}{1.72 attempts/item} \\
\addlinespace[0.5em]
\multicolumn{3}{l}{\textsc{Success Rates by Attempt Number}} \\
\quad Caption generation & \multicolumn{2}{r}{90.1\% / 8.3\% / 1.6\%} \\   
\quad Question generation & \multicolumn{2}{r}{82.7\% / 13.5\% / 3.9\%} \\  
\quad Answer generation & \multicolumn{2}{r}{71.0\% / 21.4\% / 7.6\%} \\   
\bottomrule
\end{tabular}
\label{tab:dataset_stats}
\end{table}

% ===========================================================================
\subsection{Evaluation Metrics}

To comprehensively evaluate model performance, we focus on two primary aspects: the similarity between model-generated answers and human-annotated ground truth, and the compatibility between generated responses and visual content.

For image captioning, we adopt established metrics including BLEU, METEOR, ROUGE, CIDEr~\cite{vedantam2015cider}, and SPICE~\cite{anderson2016spice}. CIDEr computes the average cosine similarity between candidate and reference captions, while SPICE measures precision and recall based on scene graphs. Together, these metrics provide a holistic assessment of caption accuracy, fluency, and contextual relevance.

For close-ended VQA tasks, we use accuracy scores, which measure the percentage of correct answers generated by the model. For open-ended VQA, where no  ground truth  exists, we follow a rubric-based approach, leveraging an LLM-as-judge~\cite{zheng2023judging} to score answers. Specifically, we use GPT-4.1 to evaluate responses based on the question, ground-truth answer, input image, and the model’s output. Each response is scored from 1 to 5 across multiple criteria: relevance, trustworthiness, clarity, and coherence\cite{pu2025judgeanythingmllmjudge}. The evaluation is repeated three times with a temperature of 0.7 to maximize consistency with human judgment.

This  evaluation strategy enables nuanced assessment of both factual correctness and the quality of higher-level reasoning in perception-related SOTIF scenarios.

\section{Model Benchmarking and Fine-tuning}
\label{sec:benchmark}

\subsection{Benchmarking}

For image captioning benchmarking,  we evaluate the BLIP \cite{li2022blip} and BLIP-2 models \cite{li2023blip} (pre-trained on the COCO Caption dataset \cite{chen2015microsoft}) across various model sizes. 
For VQA tasks, we benchmark a total of four open-sourced models, including LLaVA 1.5 \cite{liu2023llava}, Qwen2/2.5-VL \cite{wang2024qwen2vl,bai2025qwen2}, and InterVL3 \cite{zhu2025internvl3}.

These models are classified by architecture, model size, and quantization method, focusing on two specific variants: 2/3B and  7/8B. The 2/3B variant is particularly suitable for mobile device deployment due to its lightweight design and efficient performance. The 7/8B model is well-suited for on-vehicle computing, where high computational capability is required. By evaluating across this spectrum, we aim to understand trade-offs between accuracy, resource demands, and deployment feasibility for perception-related SOTIF applications in autonomous driving.

\subsection{Model Fine-tuning}
\label{sec:fine-tuning}

To enhance model performance, we apply Parameter-Efficient Fine-Tuning (PEFT) using Low-Rank Adaptation (LoRA) \cite{hu2021lora}. LoRA introduces low-rank matrices into each Transformer layer, allowing the model to adapt with far fewer trainable parameters while keeping the original model weights frozen.  This approach allows LLMs to be specialized for downstream tasks with substantially lower computational and hardware demands than full fine-tuning.

For image captioning, we fine-tuned BLIP and BLIP-2 models using the LAVIS framework~\cite{li-etal-2023-lavis}. For VQA, we utilized LLaVA~\cite{liu2023llava} and LLaMA-Factory~\cite{zheng2024llamafactory} frameworks to fine-tune LLaVA 1.5, Qwen2/2.5-VL, and other leading vision-language models. Hyperparameters, random seeds, and training configurations were kept consistent across experiments for fair comparison (details in S.II of the supplemental material).

\begin{table*}[h!]
    % \small
    \caption{Performance comparison of baseline models (Base), fine-tuned models (FT), and ground truth (GT) for visual question answering (VQA) and image captioning (IC) tasks. }
    \begin{center}
    \begin{tabular}{c l p{0.41\textwidth}  p{0.41\textwidth}}
    \toprule
    \textbf{Task} & \textbf{Type} & \textbf{Image 1}  & \textbf{Image 2} \\
    \midrule
     &  
    & \includegraphics[width=0.3\textwidth]{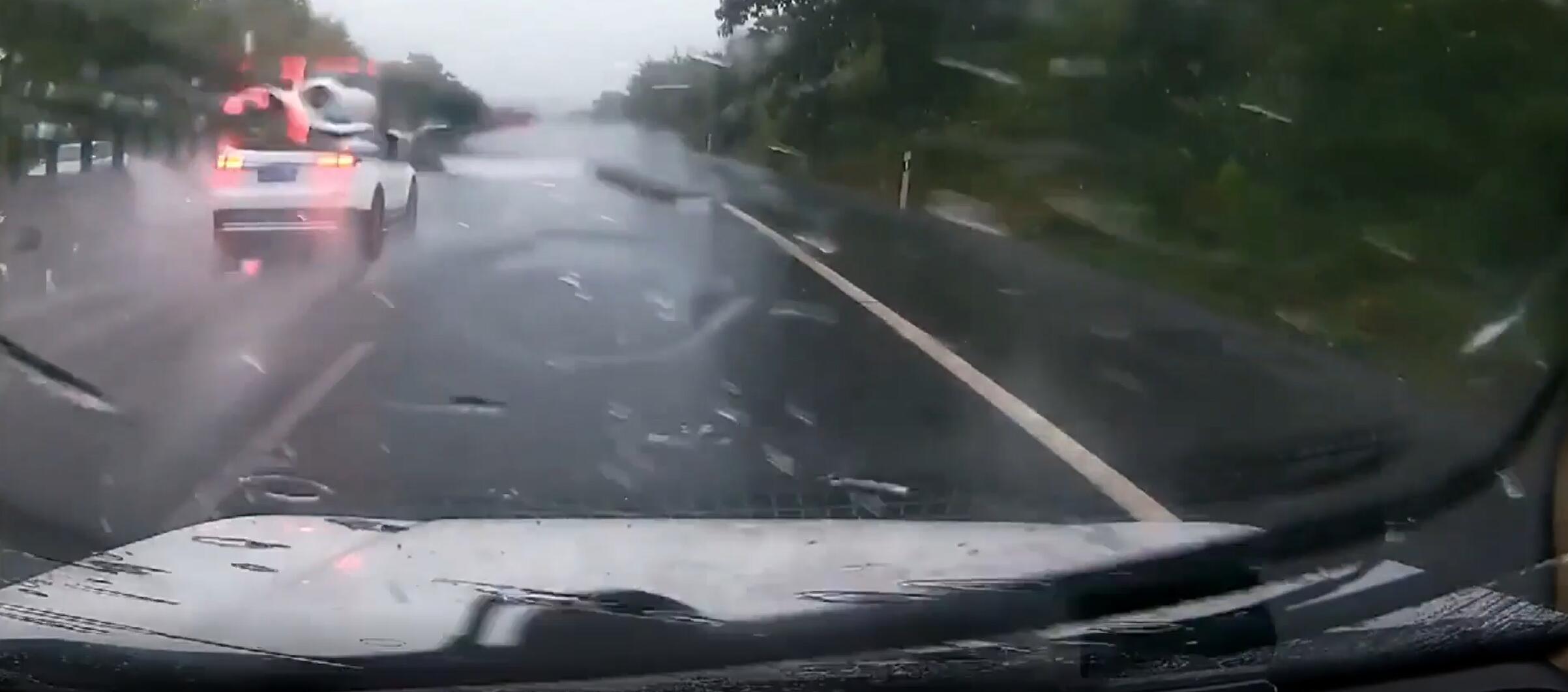} 
    % & \includegraphics[width=0.27\textwidth]
    & \includegraphics[width=0.3\textwidth]{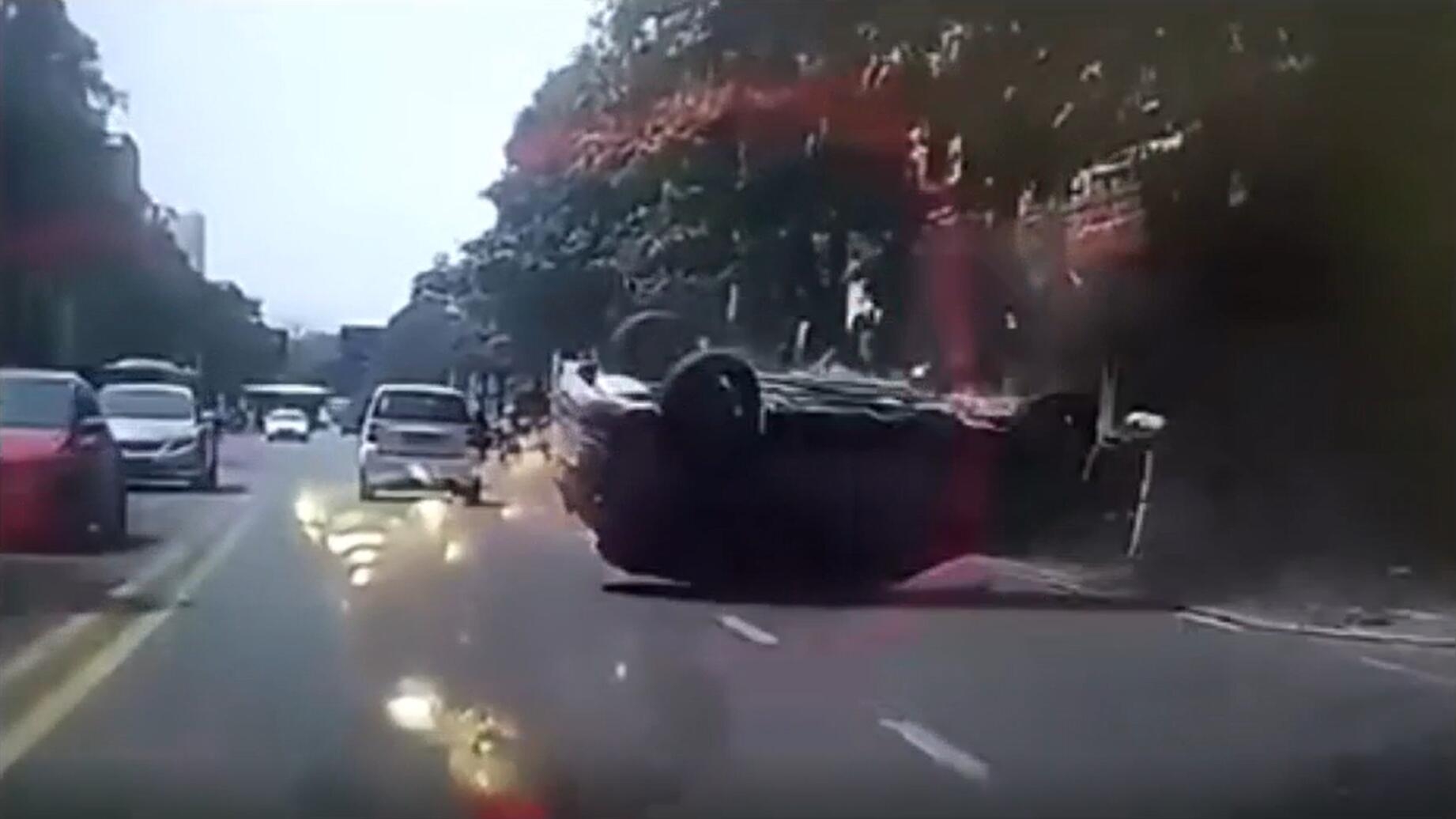} \\
    \midrule
    \textbf{IC}
    & Base:  
    & \textcolor{red}{a view of} a car driving \textcolor{red}{on a wet road} 
    % & a car driving on a foggy road 
    & a car that is upside down \textcolor{red}{on the side of the road} \\
    % \cmidrule{2-4}
    & FT: 
    & \textcolor{darkgreen}{a driver's view through a windshield on a rainy day}, with \textcolor{darkgreen}{red taillights visible} ahead 
    % & \textcolor{darkgreen}{a driver's view through a car's windshield on a foggy road}, with \textcolor{darkgreen}{low visibility and limited visibility}
    & a car \textcolor{darkgreen}{flips over on a busy road}, causing traffic congestion and hindering visibility for drivers \textcolor{darkgreen}{on a busy city street} \\
    % \cmidrule{2-4}
    & GT:  
    & View from a vehicle's dashboard camera showing a rainy and blurred roadway scene with visible windscreen wiper, a truck ahead, and blurry tail lights 
    % & Intense fog conditions on a road pose a challenge for drivers, with only faint outlines of vehicles ahead visible 
    & An overturned car in the middle of a busy street highlights potential hazards in urban traffic conditions \\
    \midrule
    \textbf{VQA}
    % & \textbf{Q1:} 
    % & \textit{Is there precipitation affecting visibility in the scenario depicted? }
    % & \textit{Is the visibility reduced in this driving scenario due to fog or mist? }
    % & \textit{Is the overturned vehicle a key object in the scene? }\\
    % \cmidrule{2-4}
    % & Base: 
    % & Yes 
    % & Yes 
    % & Yes \\
    % \cmidrule{2-4}
    % & FT: 
    % & Yes 
    % & Yes 
    % & Yes \\
    % \cmidrule{2-4}
    % & GT: 
    % & Yes 
    % & Yes 
    % & Yes \\
    % \cmidrule{2-4}
    & \textbf{Q:} 
    & \textit{What weather condition is depicted in the image that could affect perception systems in vehicles?} 
    % & \textit{What environmental factor is degrading the perception ability in this image? }
    & \textit{What perception-related SOTIF risk is present due to the vehicle's unusual position in the scene?} \\
    % \cmidrule{2-4}
    & Base:
    &  \textcolor{red}{rainy weather}, which can affect the perception systems \textcolor{red}{by reducing visibility}, making it more difficult for the systems to accurately detect and track objects.
    % &  The image shows that the environmental factor degrading perception ability is fog
    & \textcolor{red}{camera's field of view may not capture the full extent of the scene} \\
    \cmidrule{2-4}
    & FT: 
    &  The image shows \textcolor{darkgreen}{heavy rain conditions}, which can significantly degrade the performance of perception systems in vehicles by scattering and reflecting light
    % &  The perception ability is degraded by \textcolor{darkgreen}{heavy fog}, which significantly reduces visibility on the road.
    & The overturned vehicle poses a risk as it is \textcolor{darkgreen}{not commonly expected to be in that position on the road}, which can lead to \textcolor{darkgreen}{misclassification or delayed recognition} by autonomous driving systems. \\
    % \cmidrule{2-4}
    & GT: 
    & The image shows heavy rain conditions which can degrade the visibility and accuracy of vehicle perception systems by scattering light and obscuring road markings and other vehicles. 
    % & The environmental factor shown in the image is fog, which significantly reduces visibility and therefore degrades the perception ability of drivers and automotive sensors. 
    & The overturned vehicle presents a high SOTIF risk by unexpectedly altering the driving environment, which could lead to inappropriate responses from an autonomous driving system unprepared for such irregular situations. \\
    \bottomrule
    \end{tabular}
    \label{tbl:benchmark_ft_models}
    \end{center} 
\end{table*}

% \subsubsection{Quantitative Analysis}

% The quantitative results, shown in Tables \ref{table:ic_bench} and \ref{table:vqa_bench}, demonstrate that fine-tuned models consistently outperform their baseline counterparts across all evaluation metrics.

\begin{table*}[t]
% \normalsize
\centering
\caption{Image Captioning Scores}
\begin{tabular}{lccccccccccccccccc}
\toprule
\multirow{2}{*}{\textbf{Method}} & \multirow{2}{*}{\textbf{Size}} & \multicolumn{2}{c}{\textbf{BLEU-4}} & \multicolumn{2}{c}{\textbf{METEOR}} & \multicolumn{2}{c}{\textbf{ROUGE-L}} & \multicolumn{2}{c}{\textbf{CIDEr}} & \multicolumn{2}{c}{\textbf{SPICE}} \\
& & Base & FT   &  Base & FT  &  Base & FT  &  Base & FT &   Base & FT\\
\midrule
\multirow{2}{*}{BLIP} & Small &  0.082 & 0.217 & 0.134 & 0.208 & 0.315 & 0.414 & 0.138 & 0.411 & 0.086 & 0.143 \\
& Large &  0.083 & 0.206 & 0.137 & 0.210 & 0.324 & 0.399 & 0.140 & 0.410 & 0.091 & 0.145 \\
\greyrule
\multirow{2}{*}{BLIP-2} & 2.7B &  0.095 & 0.231 & 0.141 & 0.213 & 0.336 & 0.412 & 0.161 & 0.469 & 0.093 & 0.143 \\
& 6.7B &  0.106 & \textbf{0.248} & 0.146 & \textbf{0.225} & 0.338 & \textbf{0.421} & 0.176 & \textbf{0.536} & 0.096 & \textbf{0.155} \\
\bottomrule
\end{tabular}
\label{table:ic_bench}
\end{table*}

\begin{table*}[h]
% \small
\centering
\caption{VQA Scores Comparison: Baseline vs. Fine-tuned Models}

\begin{tabular}{lc  cc   c c c c c c c c c c}
\toprule
\textbf{Model} & \textbf{Size} & \multicolumn{2}{c}{\textbf{Accuracy (\%)}}    &  \multicolumn{2}{c}{\textbf{Relevance}} &  \multicolumn{2}{c}{\textbf{Trustworthiness}}  &  \multicolumn{2}{c}{\textbf{Clarity}}  &  \multicolumn{2}{c}{\textbf{Coherence}}  &  \multicolumn{2}{c}{\textbf{Overall}} \\ 
% \midrule
% Precision (%) Recall (%) F1 (%)
 &  & Base & FT &    Base & FT  &  Base & FT  & Base & FT  &  Base & FT  &  Base & FT \\
\midrule
% \multicolumn{11}{c}{\textbf{LLaVA Models}} \\
% \midrule
LLaVA 1.5  & 7B   & 57.51 & 64.06    & 3.88	& 4.01	&4.04	&4.44&	4.69	&4.85&	4.68&	4.89	&	4.32 &	4.54
\\
 \greyrule

Qwen2-VL  & 2B  & 55.84  & 61.01       & 3.70&	3.94&	3.67&	4.30	&4.33&	4.74&	4.38	&4.80&4.02 &4.45
\\
 % & & 8 bit & 19.61 & 42.85 & 25.61 & 38.14 & 5.54 & 11.67 & 14.52 & 28.94 \\ \\
 & 7B   & 56.89 & 62.13   &  4.24	&\textbf{4.14}&	4.32&	4.51	&4.81&	\textbf{4.90}&	4.79&	4.90& 4.54 &4.61
 \\
 \greyrule
Qwen2.5-VL  & 3B  & 58.33&	61.75&	4.09&	4.04&	4.05	&4.39&	4.66	&4.82	&4.70&	4.86& 4.37&  4.53
\\
& 7B  & 54.10&	60.12&	\textbf{4.57}&	4.13&	4.66&	4.51&	4.88&	\textbf{4.90}&	4.92&	\textbf{4.92}&	4.76  & 4.61  \\ \greyrule
 InternVL3  &  1B & 55.68	& 63.53	  & 3.44	&3.70	&3.44	&4.13&	4.23&	4.67&	4.23	&4.70&	3.84  & 4.30\\
 &  2B & 58.01&	\textbf{64.84}& 4.22&	3.92&	4.21&	4.34&	4.74&	4.80&	4.76	&4.82	& 4.48 &4.47  \\
  &  8B & \textbf{59.12}	& 64.49	&  4.56&	4.12&	\textbf{4.70}&	\textbf{4.53}&	\textbf{4.94}	&\textbf{4.90}&	\textbf{4.96}	&\textbf{4.92}  & \textbf{4.79} & \textbf{4.62} \\
% \greyrule
\bottomrule
\end{tabular}
\label{table:vqa_bench}
\end{table*}

% To ensure the reproducibility of our results, we maintain consistent settings across fine-tuning and evaluation phases. We use identical random seeds and hyperparameters, such as learning rate (LR), temperature, for all models. Further details on the hyperparameters and training configurations are provided in S.II of the supplemental material.

\subsection{Results and Analysis}

\label{sec:bench-result}

\subsubsection{Qualitative Analysis}

Table~\ref{tbl:benchmark_ft_models} presents representative examples illustrating how fine-tuning improves model performance on both image captioning and VQA tasks. Baseline captioning models tend to generate short, generic descriptions (e.g., “a car driving on a wet road”) that miss key scene details. After fine-tuning, models produce richer, context-aware captions such as “a driver’s view through a windshield on a rainy day, with red taillights visible ahead,” capturing both environmental context and potential hazards.

For VQA, baseline models succeed on close-ended questions but often fail to provide detailed, situation-specific reasoning. Fine-tuned models demonstrate better understanding of complex queries. For example, when prompted about rain-related perception issues, the fine-tuned model  points out specific SOTIF risk factors like reflections and distortions caused by rain that can degrade sensor performance. Similarly, in rare scenarios like an overturned vehicle, the fine-tuned model recognizes that the unusual positioning blocks the driver's field of view and may cause misclassification. In contrast, the baseline model response lacks such depth.

\subsubsection{Quantitative Analysis}

Fine-tuning improves performance across all evaluation metrics.  For image captioning, the fine-tuned BLIP-2 6.7B model obtained the best results, achieving ROUGE-L, CIDEr, and SPICE scores of 0.421, 0.536, and 0.155, respectively. This marks an increase of 24.56\%, 204.54\%, and 61.46\% compared to the baseline version. Similar improvements were observed in BLEU-4 and METEOR scores, indicating models can generate more accurate and contextually relevant descriptions of complex scenes.  Notably, the fine-tuned BLIP-2 2.7B model achieved comparable performance.

For the VQA task, larger models generally outperform smaller ones. The InternVL3-8B model achieved the highest baseline accuracy of 59.1\%. Fine-tuning significantly improved accuracy. For instance, the InternVL3-2B model increased from 58.0\% to 64.8\%, and the InternVL3-1B  improved from 55.7\% to 63.5\%.

LLM-as-judge metrics also show notable gains after fine-tuning, especially for smaller models. In detail,  InternVL3-1B improves  in relevance (from 3.44 to 3.70), trustworthiness (from 3.44 to 4.13), clarity (from 4.23 to 4.67), and coherence (from 4.23 to 4.70). Larger models like Qwen2.5-VL-7B and InternVL3-8B show less improvement in these metrics. Overall scores agree with this trend. For example, the InternVL3-1B model's overall score improves from 3.84 to 4.30, while the InternVL3-2B and InternVL3-8B models' overall scores remain or decrease slightly after fine-tuning.

This suggests that larger models already possess general reasoning capabilities but still require fine-tuning for domain-specific knowledge. Across all models,  relevance scores improved less than other metrics, indicating that baseline models can identify general risk factors but lack the detailed safety knowledge required.

In summary, fine-tuning on the DriveSOTIF dataset enhances both image captioning and VQA. Fine-tuned models generate more detailed, context-aware captions and deliver deeper, scenario-specific reasoning aligned with SOTIF-relevant safety assessments.

\section{Experiments}

% \subsection{MVSL Shuttle Bus}
\label{sec:exp}

\subsection{Model Deployment and Real-time Performance Evaluation}
MLLMs can be deployed in a range of environments, including local devices, cloud platforms, and edge computing systems such as multi-access edge computing (MEC). Popular deployment frameworks include HuggingFace, TensorRT-based solutions \cite{tensorrt_LLM}, and vLLM \cite{kwon2023efficient}. In this study, we adopt vLLM for its efficiency and support for advanced scheduling strategies.

The deployment setup follows a simplified API request-response architecture: input images and text are sent to the deployed model via API calls, and the model returns the corresponding output. For this work, compact, fine-tuned MLLMs were used to ensure fast and resource-efficient inference. These models were merged with their respective adaptive weights and deployed on two platforms: NVIDIA Jetson Orin and a consumer-grade RTX 3090 GPU.

Two inference strategies were considered. Sequential inference processes inputs one at a time as they arrive. This provides deterministic, per-sample latencies that are critical for real-time applications. Continuous batching dynamically aggregates requests and schedules token-level processing across partially completed sequences. In this research, we utilize vLLM to implement continuous batching with the goal to maximize throughput while maintaining low latency \cite{kwon2023efficient}.

\subsubsection{Sequential Processing}

%  To evaluate this mode, we test all models on 111 images from the test split of the DriveSOTIF dataset. Each image is paired with the following input prompt:
We tested this mode using 111 images from DriveSOTIF, each paired with a prompt for SOTIF risk analysis. Generation time  for each image were recorded and presented in Table~\ref{tab:multi_hardware_performance}. GPU VRAM usage was measured via HuggingFace APIs due to vLLM’s memory pre-allocation behavior.

\begin{table*}
  \centering
  \begin{threeparttable}[b]
\caption{Real-time performance across hardware platforms and quantization levels.}
\label{tab:multi_hardware_performance}
   \centering
\begin{tabular}{lllcrrrrrr}
\toprule
& & & & \multicolumn{3}{c}{\textbf{RTX 3090}}  & \multicolumn{3}{c}{\textbf{Jetson AGX Orin\tnote{1}}} \\
% \cmidrule(lr){4-6} \cmidrule(lr){7-9} 
\cmidrule(lr){5-7} \cmidrule(lr){8-10}
\textbf{Model} & \textbf{Size} &  \textbf{Quant} & \textbf{RAM Usage} & \textbf{Mean} & \textbf{Median} & \textbf{Std} & \textbf{Mean} & \textbf{Median} & \textbf{Std}  \\
& & & \textbf{[GB]} &  \textbf{[s]} & \textbf{[s]} & \textbf{[s]} & \textbf{[s]} & \textbf{[s]} & \textbf{[s]} \\
\midrule
\multirow{3}{*}{Qwen2-VL} & \multirow{3}{*}{2B}
& - & 4.74 & 0.82 & 0.80 & 0.11 & 3.70 & 3.60 &0.27 \\
& & FP8 & 2.91 & 0.74 & 0.73 & 0.03 & 2.93 & 2.89 & 0.18\\
& & 4bit  & 2.05 & 0.87 & 0.86 & 0.04 & 3.56 & 3.49 & 0.23\\
\greyrule
% Qwen2.5-VL
\multirow{2}{*}{Qwen2.5-VL}
& \multirow{2}{*}{3B}
 & - & 7.64 & 1.39 & 1.38 & 0.02 & 6.86 & 6.75 & 0.36 \\
% & & 8bit &  &  &  &  &  &  &  & &  &   \\
&  & 4bit  & 2.90 &1.01 & 1.00 & 0.03  & 5.63 & 5.61 &0.21 \\
\greyrule
% \multirow{3}{*}{SmolVLM2}
% & 0.26B
% & - &  0.51& 0.42 &0.43 & 0.07 & - & - & -\\
% % & 8-bit Quant & 3.0 &  &  &  &  &  &  & 1.44 & 1.40 & 0.18 \\
% % & 4bit  & 2.0 &  &  &  &  &  &  & 0.59 & 0.57 & 0.08 \\
%  & 0.5B
% & - & 1.00 & 0.54 &0.53 & 0.14 & - & - & -\\
% % & 8-bit Quant & 3.0 &  &  &  &  &  &  & 1.44 & 1.40 & 0.18 \\
% % & 4bit  & 2.0 &  &  &  &  &  &  & 0.59 & 0.57 & 0.08 \\
% % \greyrule
% &  2.2B
% & - & 4.27 & 1.08 &1.12 & 0.09 & - & - & -\\
% % & 8-bit Quant & 3.0 &  &  &  &  &  &  & 1.44 & 1.40 & 0.18 \\
% % & 4bit  & 2.0 &  &  &  &  &  &  & 0.59 & 0.57 & 0.08 \\
% \greyrule
\multirow{2}{*}{InternVL3\tnote{2}} & 1B    
& - & 1.99 & 0.59 & 0.57 & 0.14 & 2.16  & 2.14 & 0.26\\
& 2B & - & 4.14 & 1.09 & 1.07 & 0.16 & 4.21 & 4.14 &0.40 \\
\bottomrule
\end{tabular} 
     \begin{tablenotes}

       \item [1] We utilized vLLM docker for this testing. Orin is set to MAXN. We encountered system throttling warnings due to over-current. 
       \item [2] LMDeploy and vLLM are used for InternVL3 deployment.
     \end{tablenotes}
  \end{threeparttable}
\end{table*}

The peak GPU memory usage and per-image processing time were recorded, and the results are presented in Table~\ref{tab:multi_hardware_performance}. The analysis shows that quantization reduces GPU memory consumption and improves inference speed. For Qwen2-VL (2B), 4-bit quantization achieves the lowest memory footprint (2.05 GB), while FP8 delivers the fastest inference on both the RTX 3090 and Jetson Orin, with an average latency reduction of 14.7\% compared to the full-precision model. A similar trend is observed for Qwen2.5-VL (3B), where 4-bit quantization significantly reduces memory usage (from 7.64 GB to 2.90 GB) and lowers mean inference time from 6.86 s to 5.63 s on Jetson Orin.

While 4-bit quantization is more memory-efficient, it runs slightly slower than FP8 on both platforms (3.56 s vs. 2.93 s on Orin). This is likely due to current limitations in vLLM’s integration with bitsandbytes, which lacks fully optimized GPU kernels for INT4 inference and introduces additional overhead through de-quantization or fallback routines.

Among all evaluated models, InternVL3-1B consistently delivers the most efficient performance, with the lowest latency (0.59 s on RTX 3090, 2.16 s on Jetson Orin) and the smallest memory footprint (1.99 GB). These characteristics make InternVL3-1B particularly well-suited for real-time inference on resource-constrained platforms. In contrast, larger models such as InternVL3-2B demand more memory and exhibit higher latency, limiting their practicality on edge devices.

\subsubsection{Continuous Processing for high-speed driving}

\begin{figure}[t]
    \centering
    \includegraphics[width=0.9\linewidth]{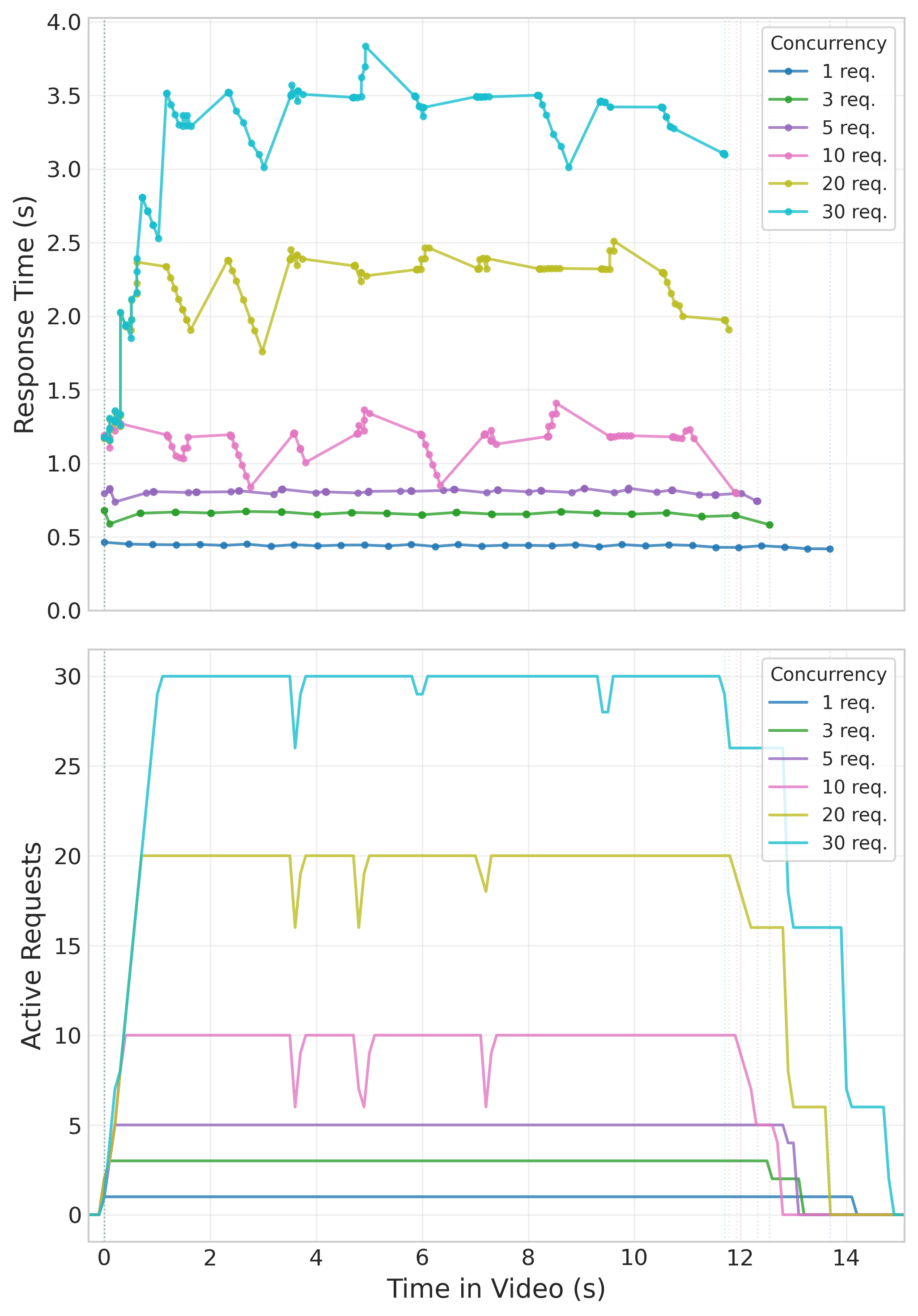}
    \caption{
Average generation time per image in continuous inference mode.}
    \label{fig:continuous_inference}
\end{figure}

To evaluate continuous inference, we replayed four 30 Hz video streams of high-speed driving scenarios, triggering model inference on every frame. Each frame generates a request to the InternVL3-1B vision-language model, which is deployed on an RTX 3090 with continuous batching enabled \footnote{Please refer to Section V of the supplemental material for results on RTX4090 and dual RTX4090 setups}. The system supports multiple requests in parallel, and both queuing and auto-timeout mechanisms are in place: when more requests arrive than the hardware can process immediately, excess requests are placed in a first-in, first-out (FIFO) queue, and any request waiting or processing beyond a set timeout threshold is automatically dropped. We varied the maximum number of in-flight (active) requests from 1 up to 30 (concurrency). For each setting, we recorded the API response time and the number of active requests throughout the experiment (see Figure~\ref{fig:continuous_inference}).

At low concurrency levels (1, 3, 5), response times remained stable and low, averaging around 0.5 to 0.7 seconds per frame. As concurrency increased past 10, response times began to rise rapidly, reaching an average of 3.5 seconds per frame at the highest level tested (30). This sharp jump signals that the model and GPU are saturated, leading to longer waits and increased resource contention.

\begin{figure}[t]
    \centering
    \includegraphics[width=0.9\linewidth]{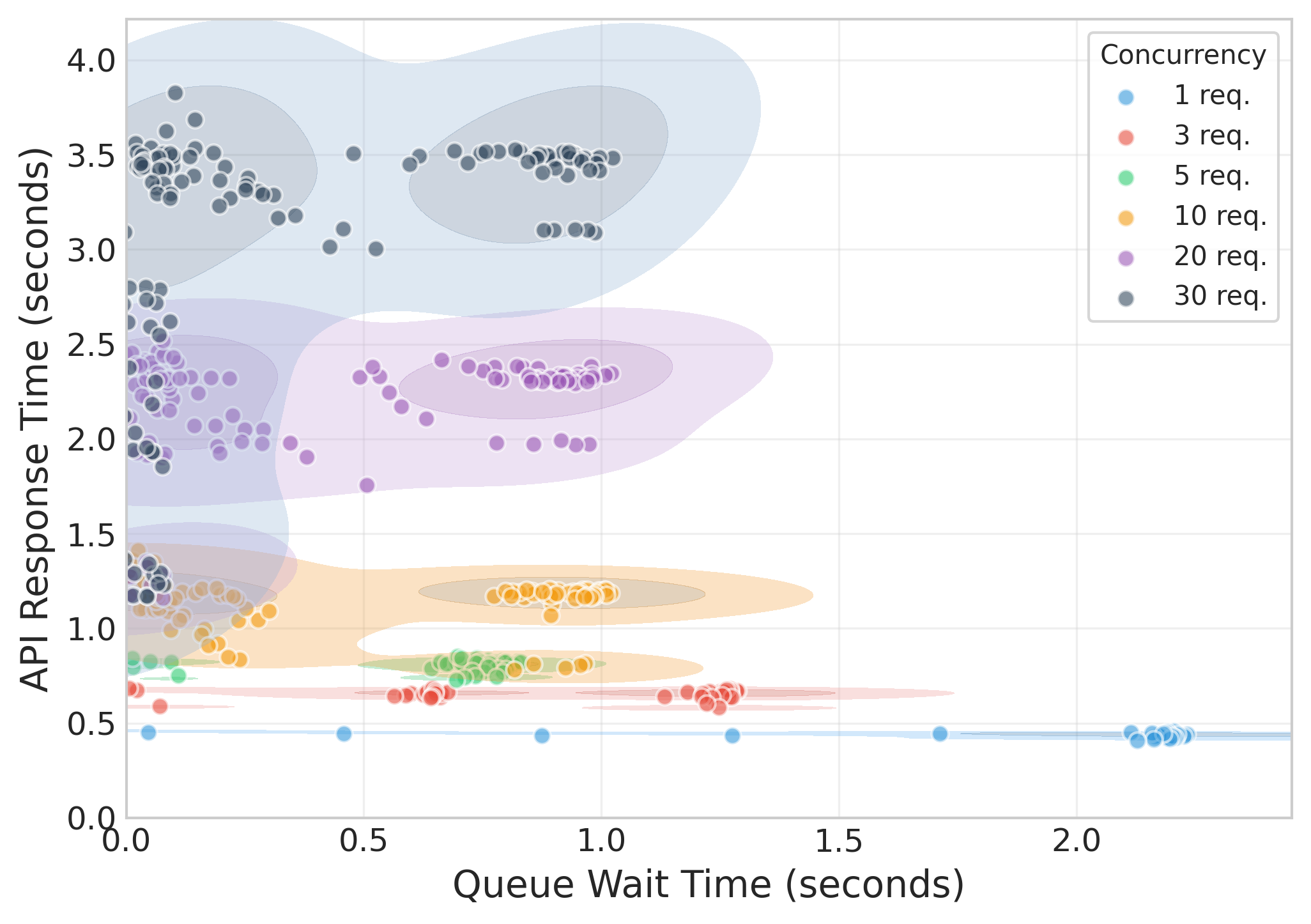}
    \caption{
Tradeoff between API response time and queue waiting time in continuous inference mode.}
    \label{fig:continuous_inference_tradeoff}
\end{figure}

We visualized these dynamics by jointly plotting queue wait times and API response times for each concurrency setting (Figure~\ref{fig:continuous_inference_tradeoff}). At low concurrency,  the system processes requests quickly once they reach the model, but incoming requests rapidly accumulate in the queue, leading to prolonged wait times and, frequently, discarded requests due to timeouts. As concurrency increases, response times become more variable and overall latency rises, but queue wait times decrease since more requests are processed in parallel.

This illustrates a fundamental tradeoff: high concurrency increases per-request processing times but keeps queue wait times low, while low concurrency reduces processing latency but leads to longer queues and more dropped requests under heavy load. In practice, we recommend using low concurrency for critical tasks that require rapid responses and higher concurrency for less time-sensitive workloads. This strategy matches the dual-mode architecture or fast-slow mechanism seen in recent vision language model (VLM) and visual language action (VLA) model research \cite{xu2025humancentricautonomousdrivingfastslow,tang2025vlmplannerintegratingvisuallanguage}. In these systems, real-time software handles high-speed, routine processing (fast mode), and the VLM or VLA model is triggered when needed to provide semantic reasoning (slow mode).

    % \label{fig:continuous_inference}

\subsection{Real World Data Evaluation}
For real-world evaluation, as shown in Table~\ref{tbl:exp_result}, we evaluated the fine-tuned models using data collected in Waterloo, Ontario, Canada, along with proprietary dash-cam footage from China. This evaluation is conducted through a ROS code that queries  the MLLM API with random questions from a list, along with image data from collected data in the form of ROSBag or video.

% \subsection{Waterloo, Ontario, Canada}
In detail, we used image and video data from the University of Waterloo's WATonoBus project \cite{cui2023drivellm} and the Canadian Adverse Driving Conditions (CADC) dataset \cite{pitropov2021canadian}. The WATonoBus is an autonomous shuttle operating along a 2.7-kilometer, two-lane University Ring Road that encircles the campus. The CADC dataset was gathered by an autonomous vehicle platform navigating various areas around Waterloo. 

The first scenario features a road maintenance crew working on a snowy road inside the university campus with a snowplow in operation. The snowplow, being a rare object, along with the workers and traffic cones, pose challenges for perception systems. 
The second scenario involves a truck and trailer preparing to merge into the ego vehicle's lane on a snowy, slippery residential road, where visibility is reduced due to the adverse weather conditions. 
These two scenarios effectively demonstrate the SOTIF challenges faced by autonomous vehicles in both public roads and university campus environments.

In China, we evaluated the models on two challenging scenarios.
% \textbf{1) Hidden child in a flattened box:}
The first scenario involved a child hidden in a flattened box lying on the ground, with only a small portion of his foot  visible. This scenario is particularly difficult, as the hazard is subtle and easily missed by both human drivers and autonomous systems. In this case, the human driver successfully stopped in time, avoiding an accident. Our aim is to examine whether the fine-tuned MLLM can detect these tiny yet critical visual cues to prevent such accidents.
% \textbf{2) Night driving on a wet road:}
The second scenario took place at night, where the ego vehicle was driving on a wet, reflective road surface due to rain. The glare from the headlights of an oncoming vehicle further complicated the situation, making it difficult to perceive the road and surrounding environment clearly.   In this scenario, an electric bicycle crossed the road median unexpectedly and collided with the ego vehicle, resulting in an accident.

\begin{table*}[t]
    % \centering
    \centering
    % \small
    % % \small
    \caption{Results of Real-World Evaluation}

        \begin{tabular}{m{0.43\textwidth}  m{0.43\textwidth}}
        \toprule
% \multicolumn{2}{c}{\textbf{Ontario, Canada} } \\ \midrule
\textbf{Waterloo} - Snow Plowing, University of Waterloo campus & \textbf{Waterloo} - Merging Scenario, city of Waterloo \\        \midrule
        \begin{center}
                \includegraphics[height=4cm]{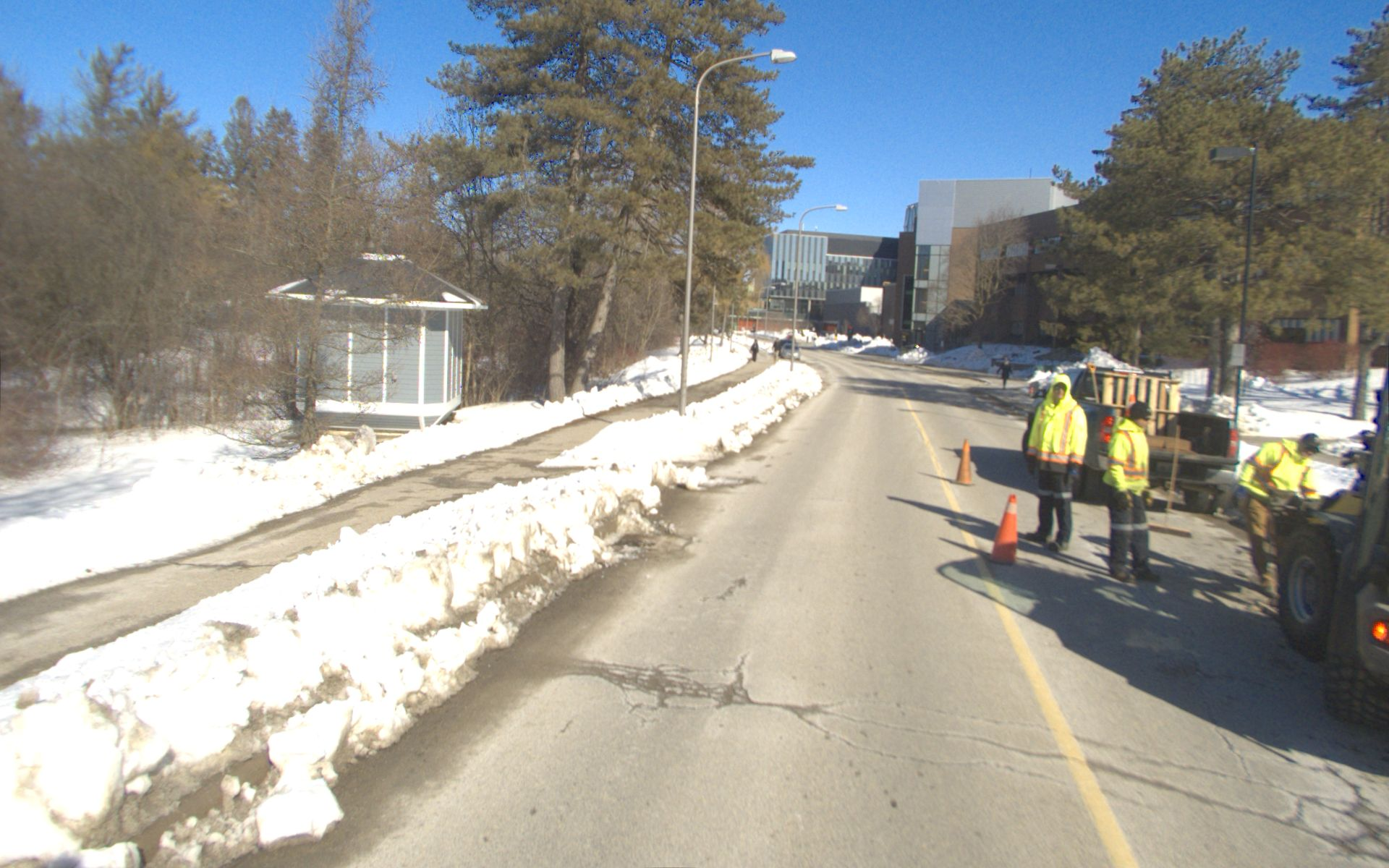} \end{center}

& 
      \begin{center}
 \includegraphics[height=4cm]{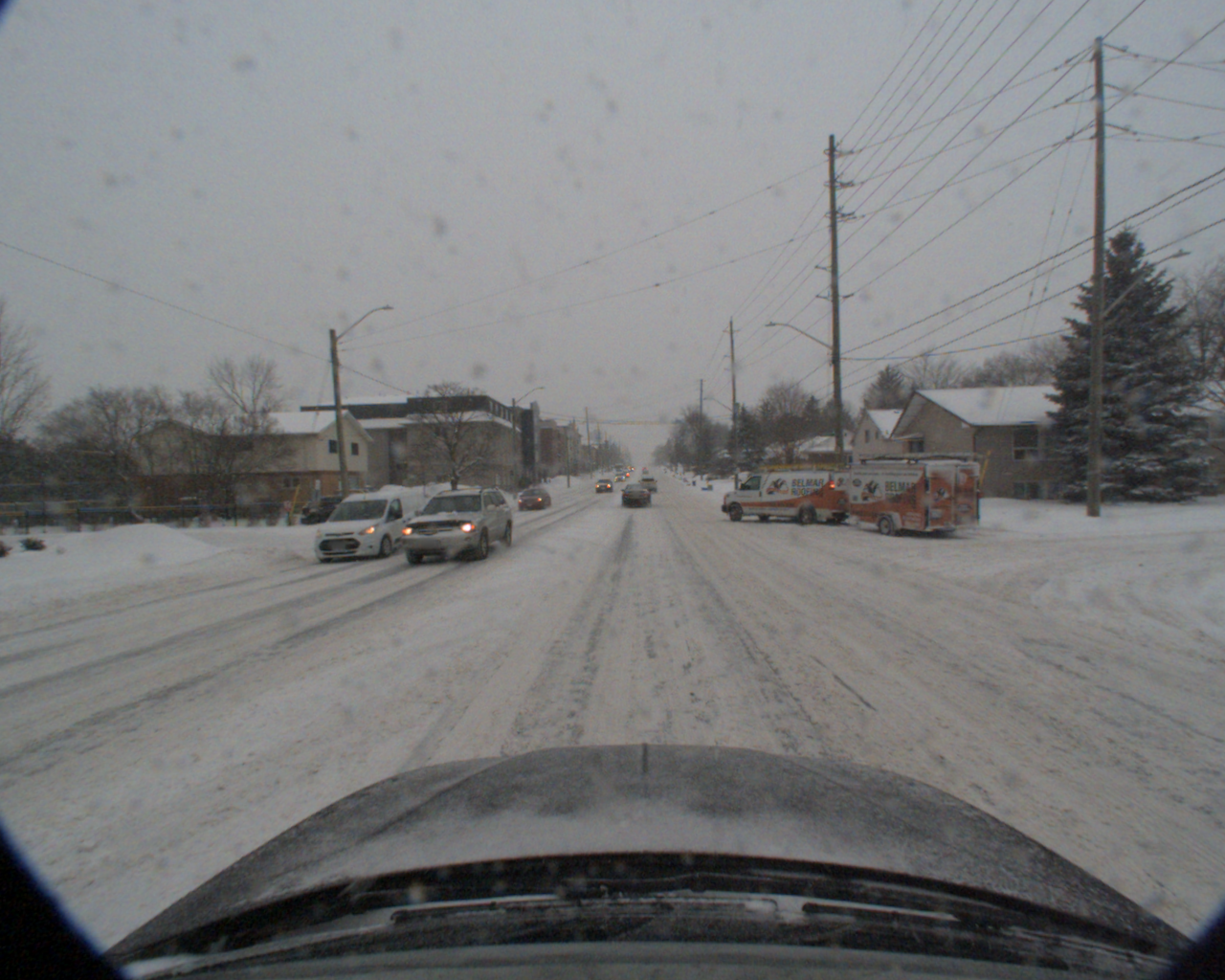}  \end{center}
\\
        % \midrule
        \greyrule
        \textbf{Q}: What are the perception-related SOTIF risks evident in this image? & 
        \textbf{Q}: What is the cause of the perception-related SOTIF risks in this image? 
        \\
        \textbf{A}:  The risks include potential misidentification of road boundaries due to snow, reduced contrast between the road and its surroundings, and the \textbf{ presence of workers and equipment} which could be obscured by snow or poorly illuminated.
 &          \textbf{A}:  The perception-related SOTIF risk in this image is caused by heavy snowfall, which can \textbf{reduce visibility and obscure road markings, signs, and other vehicles}. \\
        % \bottomrule \\

\multicolumn{2}{c}{\textbf{}} \\
        \toprule

% \multicolumn{2}{c}{\textbf{China}} \\ \midrule
\textbf{China} - Kid hiding in a box &  \textbf{China} - Strong Vehicle Front Lights, with reflective road\\
        \midrule
      \begin{center}
  \includegraphics[height=4cm]{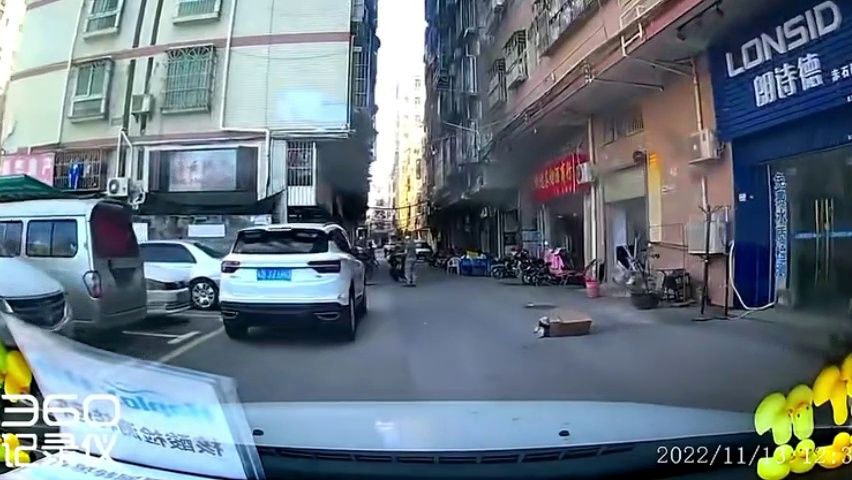}  \end{center}
& 
       \begin{center}
 \includegraphics[height=4cm]{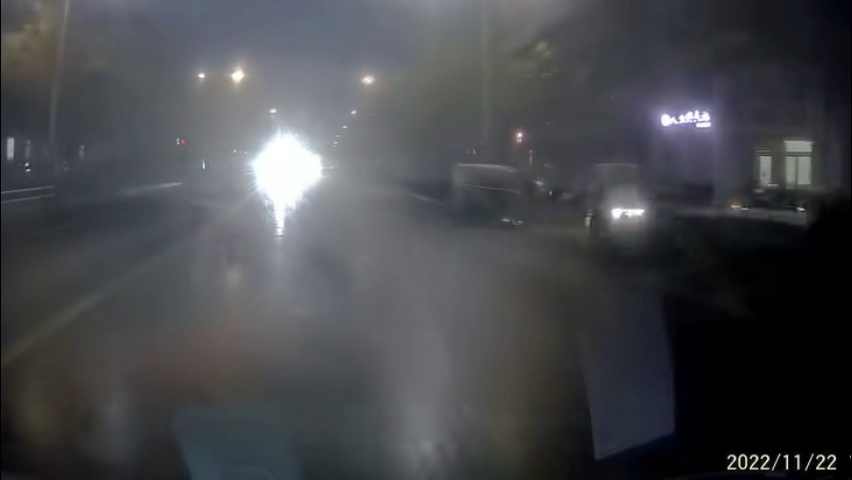}  \end{center}
\\
        \greyrule

        \textbf{Q}: What are the SOTIF risks, and what's the cause?  &          \textbf{Q}: What are the perception SOTIF risks and what's the recommended action for the ego vehicle? \\
        \textbf{A}:  The SOTIF risks include the potential for an autonomous vehicle's sensors to misinterpret the presence and position of objects on the road, such as the \textbf{boxes} and the bicycle, due to the cluttered and irregular street setup. This could lead to inappropriate vehicle responses or \textbf{failures to detect critical road users}.& 
 \textbf{A}:  The perception SOTIF risks include \textbf{failure to detect lane markings, nearby vehicles, or pedestrians due to reduced visibility}. The recommended action includes activating hazard lights and slowing down, and possibly engaging in manual driving if the perception system fails. \\
        \bottomrule
    % \end{minipage}
            \end{tabular}

    \label{tbl:exp_result}
\end{table*}

\subsection{Results} 
As shown in Table~\ref{tbl:exp_result}, in our experiments, fine-tuned MLLMs performed effectively in challenging environments, yet it also highlighted critical areas for enhancement. Notably, in Waterloo's snowy conditions, the MLLM successfully managed complex scenarios with obscured road markers and temporary obstacles like snowplows and maintenance crews. 
% This performance  highlights  the model's robust feature recognition capabilities under adverse weather conditions, proving its potential  in dynamic real-world applications. 

In the Strong Vehicle Front Lights scenario, despite intense glare from oncoming vehicle headlights, the MLLM successfully identified potential risks and provided reasonable recommended actions. The model recognized that the glare could significantly reduce visibility, increasing the likelihood of missing nearby vehicles or pedestrians. In response, it suggested actions that could directly address the visibility issue and enhance safety, such as slowing down, increasing following distance, or activating hazard lights.

However, the model struggled to detect subtle hazards, such as the hidden child in the China experiments.
 While the model detected unusual patterns associated with the obscured visual cues, it did not accurately capture and interpret the partially or fully obscured child. This highlights a scenario that is challenging even for human drivers. One possible reason is the current capacity of the visual encoders used in the MLLM.

\section{Discussion}

\subsection{Implications of Experimental Results}

Based on experimental results, the model demonstrates open-world generalization by correctly identifying hazards in unseen scenarios. The model also exhibits causal reasoning by linking the combination of nighttime driving, wet roads, and headlight glare to increased risks. Lastly, it shows contextual understanding by recognizing that pedestrians standing next to snowplows are maintenance workers and providing relevant safety assessments and recommendations.

The computational efficiency reported in Section V.A validates that the above capabilities can be achieved without excessive resource demands. In detail. The fine-tuned InternVL3-1B model requires only 2 GB of GPU memory and can achieve inference times as low as 0.59 seconds per image. This makes it possible to deploy the model onboard vehicles as a lightweight, near-real-time risk assessment module. Doing so removes dependence on external APIs, reduces exposure to privacy and cybersecurity risks, and eliminates the need for continuous internet connectivity.

Pairing a fine-tuned model  with a decision-making framework like DriveLLM\cite{cui2023drivellm} enables autonomous systems to incorporate SOTIF risks into real-time decision-making. This integration improves safety in complex environments and potentially prevents  fatal accidents like the child-hidden-in-the-box scenario, where conventional AD systems may fail to anticipate hidden risks. With iterative dataset expansions and human-in-the-loop annotation, the model can be continuously refined and grounded using SFT and reinforcement learning from human feedback (RLHF) \cite{dai2023safe}.  This adaptive approach enables MLLMs to expand   to  tasks like complex scenario understanding and ODD assessment.

\subsection{Domain Transfer}

Results in Table~\ref{tbl:vqa_sample} show that  proprietary LLM models can provide accurate and contextually relevant answers to complex, domain-specific SOTIF questions without specialized training. This highlights their effectiveness in zero-shot and one-shot scenarios, where little or no task-specific data is available.

Moreover, as seen in Tables \ref{tbl:benchmark_ft_models} and \ref{tbl:exp_result},  SFT processes enable baseline models to incorporate domain-specific knowledge. 
These fine-tuned models perform reasonably well when answering questions about SOTIF challenges for vehicles operating in both Waterloo, Canada, and China.  This indicates that the models can effectively transfer learned knowledge across different environmental contexts, adapting to diverse traffic scenarios, weather conditions, and regulatory landscapes.

That said, there are certain limitations to this approach. While domain transfer allows for generalization across different contexts, model performance may be less consistent in environments with extreme conditions or significantly different operational parameters. Although the models adapt well to many contexts, some scenarios may still require additional fine-tuning to maintain accuracy and reliability, especially when facing highly specific regulatory or environmental challenges.

\subsection{Hallucination}
Hallucination in LLMs refers to the generation of plausible yet factually incorrect or inconsistent responses. It represents a critical challenge for safety-critical applications like autonomous driving. Hallucination can occur at various stages, including pre-training, post-training, SFT, and inference\cite{zhang2023siren}. Within this research, we focus specifically on hallucination suppression during the SFT stage.

During dataset generation, we observed instances of object hallucination, where VLMs would inconsistently report the presence of objects across multiple queries. For example, when asked whether a specific object was present in an image, one out of three attempts indicated "yes," while the remaining attempts responded "no." This inconsistency exemplifies fact-conflicting hallucination, a critical reliability issue in VLMs where models generate contradictory responses to identical queries \cite{zhang2023siren}.

To address this challenge, we implemented a multi-stage hallucination suppression strategy during the SFT process. Our approach includes (1) a post-processing validation step using an independent LLM to cross-verify generated responses against input images, (2) consistency filtering that removes outputs failing validation checks, and (3) iterative correction of inconsistent annotations. This targeted intervention significantly reduces hallucinated content entering the fine-tuning pipeline, ensuring our dataset maintains grounding in factual, image-based annotations.

However, inference-time hallucination remains a persistent challenge, particularly due to inherent knowledge gaps and biases in pre-trained model backbones. Existing mitigation strategies present distinct trade-offs for real-time applications. Retrieval-augmented generation \cite{lewis2020retrieval}, while effective at grounding responses in factual content, introduces additional inference latency that may be incompatible with the sub-second response requirements of autonomous driving systems. Human feedback mechanisms \cite{lee2024rlaifvsrlhfscaling}, though widely adopted in industry and demonstrably effective at improving model reliability, face scalability constraints when applied to high-frequency inference scenarios typical in autonomous vehicles. At the moment, one feasible direction  is grounding model responses through a human-in-loop semi-auto feedback mechanism. Other methods, like latent space steering \cite{liu2024reducing}, can also be explored to reduce hallucination while maintaining real-time performance.

\subsection{Future of LLM in SOTIF}

As current research on LLMs in SOTIF applications remains limited, we outline several promising directions for both text-based and multimodal LLMs. 
% These applications can work together to create a more complete SOTIF assessment system.
First, a universal description system for SOTIF scenarios can be developed to provide standardized descriptions. Building on this foundation, custom-trained LLMs can generate JSON-based representations for scenario detection, communication, and risk identification. This standardization would make it easier to share and compare SOTIF scenarios across different research groups and industry applications.

These standardized descriptions would then support better testing approaches. LLMs can efficiently generate SOTIF test cases following systematic testing plans. By learning from test failures, these systems can improve subsequent tests, reducing the total number of tests needed. LLMs can also create highly targeted test cases for specific SOTIF risks, focusing on edge cases and rare conditions that are most challenging and potentially dangerous.

The testing capabilities naturally extend to evaluation tasks. Multimodal LLMs can assist in evaluating existing rule-based or learning-based SOTIF methods. By providing advanced scene understanding, MLLMs can refine manually defined evaluation algorithms and cross-check results from conventional systems. This ensures SOTIF risks are assessed from multiple angles. MLLMs can also select specific scenarios of interest based on input prompts.

Finally, multimodal LLMs can integrate data from multiple sensor types, such as LiDAR, radar, and bird's-eye-view images. By encoding these inputs into feature embeddings alongside textual prompts, LLMs could directly assess SOTIF risks at the feature level. This approach would provide more detailed risk understanding, reduce reliance on processed detection results, and potentially identify hazards earlier in the perception pipeline.

\section{Conclusion and Future Work}
% \section{Conclusion and Future Work}

In this paper, we demonstrate that domain-specific fine-tuning transforms general-purpose MLLMs into specialized models that can analyze, reason about, and react to perception-related SOTIF risks. DriveSOTIF, as the first VQA and image-captioning dataset in the SOTIF domain, enables this adaptation and provides a reproducible framework for evaluation and benchmarking.

Our results can be summarized into three key points. First,  fine-tuned MLLMs generate more detailed, context-aware captions and deliver deeper, scenario-specific reasoning aligned with SOTIF-relevant safety assessments. 
Second, these models demonstrate open-world generalization, causal reasoning, and contextual understanding  across regions, weather conditions, and traffic scenarios. Lastly, these models achieve an average inference time of 0.59 seconds per image with only 2 GB of GPU memory, making them suitable for low- to moderate-speed applications like urban navigation, automated parking, and traffic jam assistance.

Deploying these models locally on vehicle hardware enhances driving safety while preserving user privacy and removing dependence on external APIs or cloud connectivity. When integrated into existing decision-making systems, they allow autonomous vehicles to account for SOTIF risk in real time. Crucially, these models can be continuously refined via iterative dataset updates and human feedback. This allows MLLMs to improve over time and expand to more complex tasks like scenario understanding and ODD assessment.

For future work, we plan to improve real-time performance for high-speed safety-critical applications and look into hallucination mitigation during the inference stage. We will also expand the current evaluation to more recent VLM models. At the same time, unifying SOTIF risk analysis and cognition along with action generation using VLA frameworks is a promising direction, given the current trend toward end-to-end autonomous driving systems.

% \newpage
\appendices

\section*{Acknowledgment}

The authors would like to acknowledge the financial support of the Natural Sciences and Engineering Research Council of Canada (NSERC) and Vector Institute. We also acknowledge the research grant provided by OpenAI.
%  and research grant provided by OpenAI 

% Addiitonalsd

% We would thank Lamdalabs for providing GPU compute.

% Additionally, we would like to acknowledge the Researcher Access Program provided by OpenAI

\ifCLASSOPTIONcaptionsoff
  \newpage
\fi

\bibliographystyle{IEEEtran}
\bibliography{ref}
  % \printbibliography

% % \vspace{-0.4cm}

\begin{IEEEbiography}[{\includegraphics[width=1in,height=1.25in,clip,keepaspectratio]{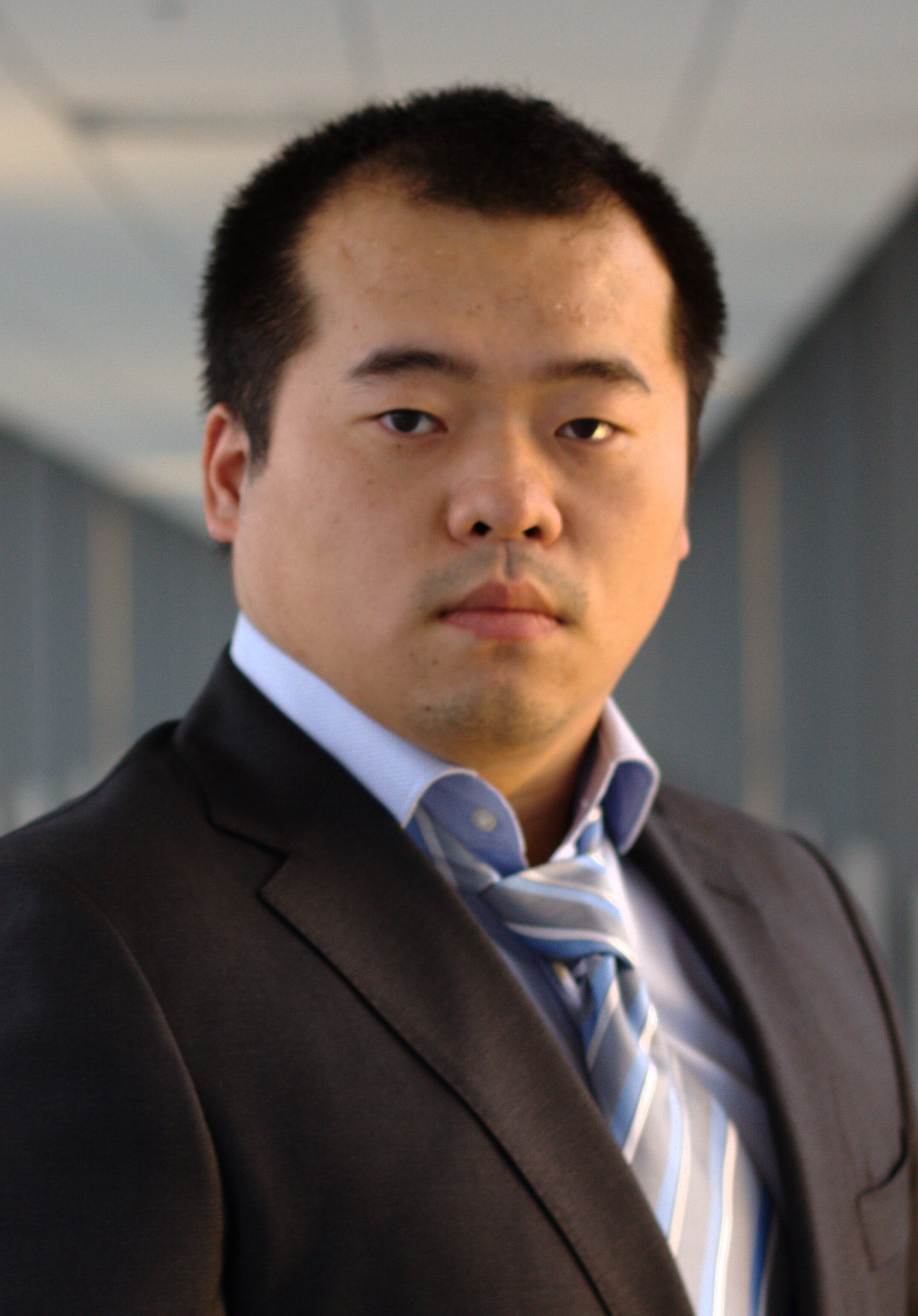}}]{Shucheng Huang} received the B.S. degree in mechanical engineering from Pennsylvania State University, State College, USA, in 2018, and the MASc degree in mechanical and mechatronics engineering from the University of Waterloo, Waterloo, Canada, in 2020. He is currently a Ph.D. candidate at the University of Waterloo Mechatronic Vehicle Systems (MVS) Lab and CompLING Lab and a graduate student member at the Vector Institute. 
His research interests include the application of large language models and machine learning in autonomous driving and energy domains. 
\end{IEEEbiography}

% \vspace{-0.4cm}

\begin{IEEEbiography}[{\includegraphics[width=1in,height=1.25in,clip,keepaspectratio]{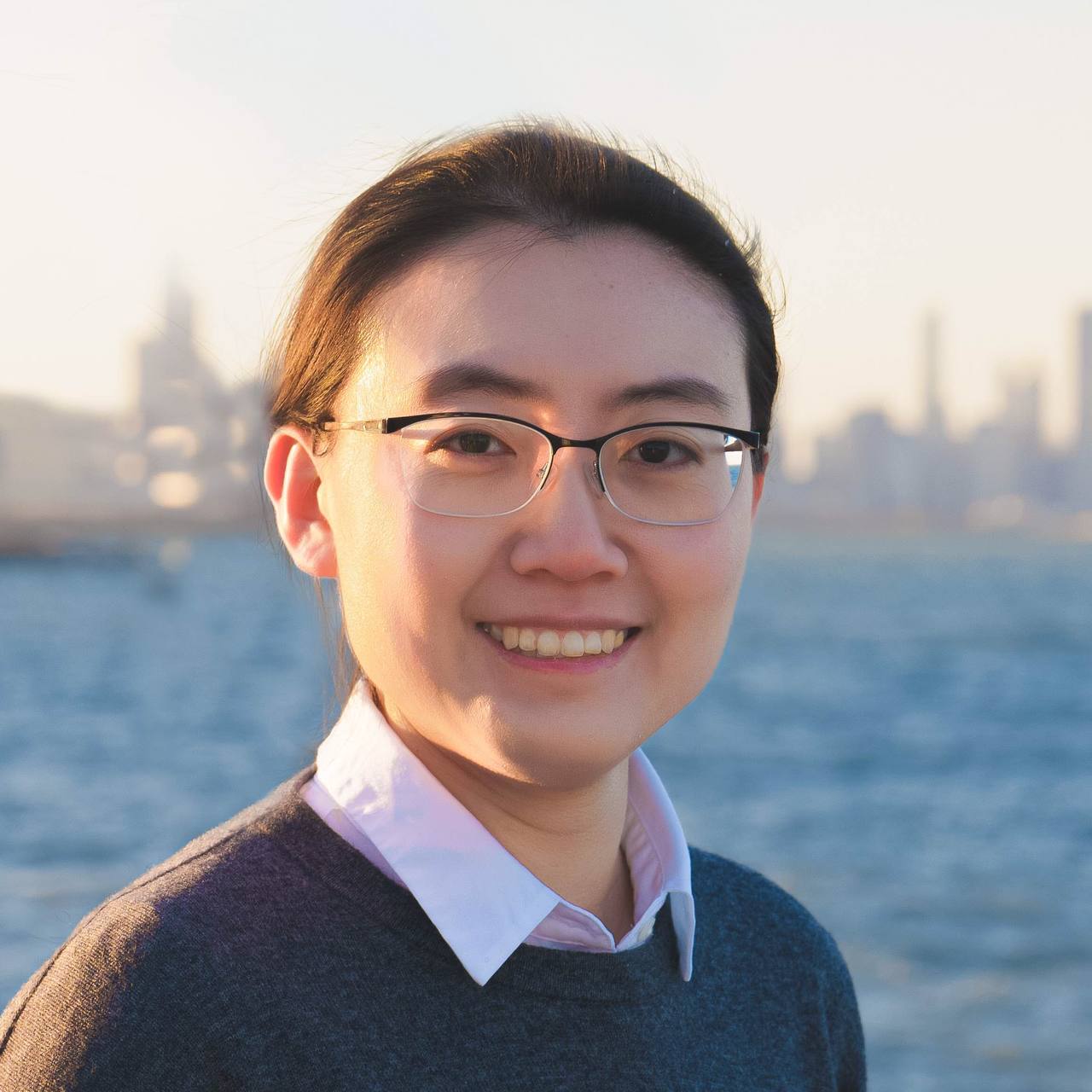}}]{Freda Shi} 
received her Ph.D. degree in Computer Science from the Toyota Technological Institute at Chicago in 2024. She is currently an Assistant Professor in the David R. Cheriton School of Computer Science at the University of Waterloo and a Faculty Member at the Vector Institute. She completed her Bachelor's degree in Intelligence Science and Technology with a minor in Sociology from Peking University in 2018. Her research interests include natural language processing, computational linguistics, machine learning, and the application of artificial intelligence in real-world scenarios.

\end{IEEEbiography}
% \vspace{-0.4cm}

\begin{IEEEbiography}[{\includegraphics[width=1in,height=1.25in,clip,keepaspectratio]{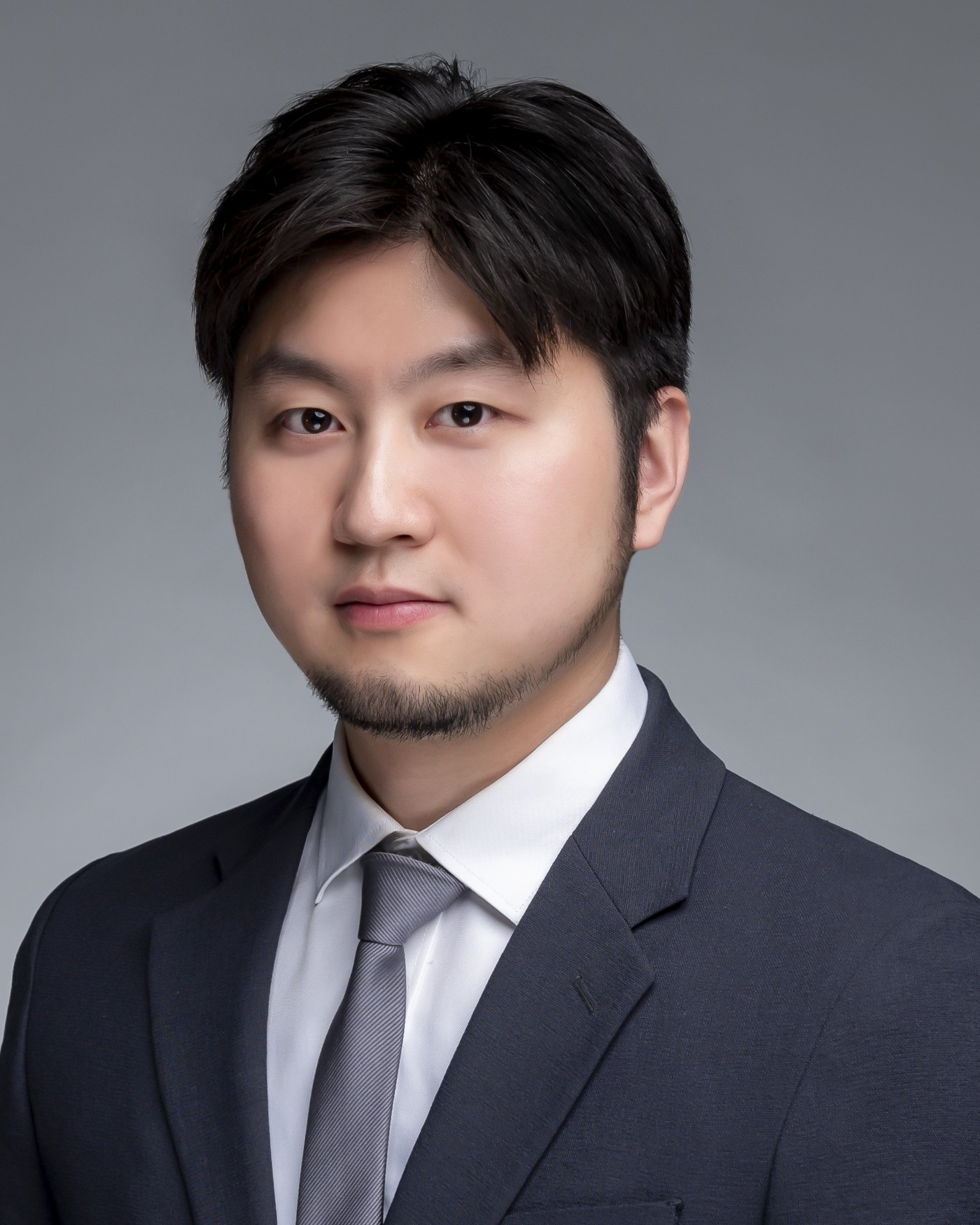}}]{Chen Sun}
  received the Ph.D. degree in Mechanical \& Mechatronics Engineering from the University of
  Waterloo, ON, Canada in 2022, M.A.Sc degree in Electrical \& Computer Engineering from the University of Toronto, ON, Canada in 2017, and B.Eng. degree in automation from the University of Electronic Science and Technology of China, Chengdu, China, in 2014. He is currently an Assistant Professor with the Department of Data and Systems Engineering, University of Hong Kong. His research interests include field robotics, safe and trustworthy autonomous driving and in general human-CPS autonomy.
\end{IEEEbiography}% \vspace{-0.4cm}

\begin{IEEEbiography}[{\includegraphics[width=1in,height=1.25in,clip,keepaspectratio]{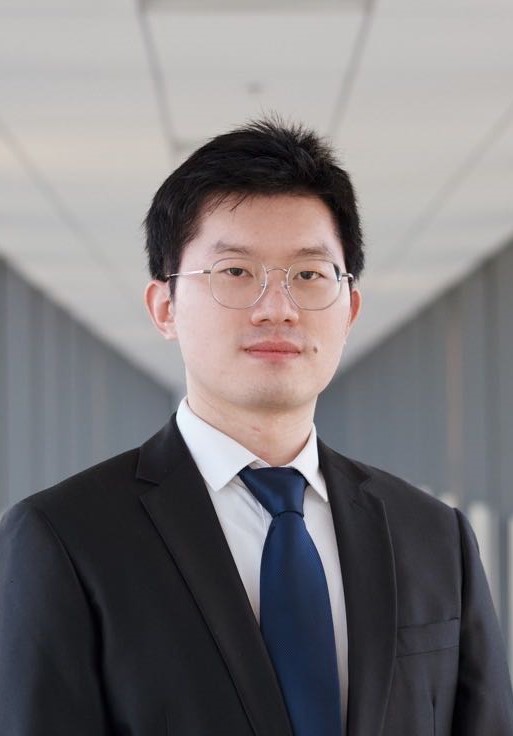}}]{Jiaming Zhong} received Ph.D. degree at the University of Waterloo in 2025 from the Mechatronic Vehicle Systems (MVS) Lab. He received the B.S. and the MASc degrees in mechanical engineering from Beijing Institute of Technology, China, in 2014 and 2017. His research interests include learning-based planning and control, multi-agent theory, and autonomous driving. 
\end{IEEEbiography}
% \vspace{-0.4cm}

\begin{IEEEbiography}[{\includegraphics[width=1in,height=1.25in,clip,keepaspectratio]{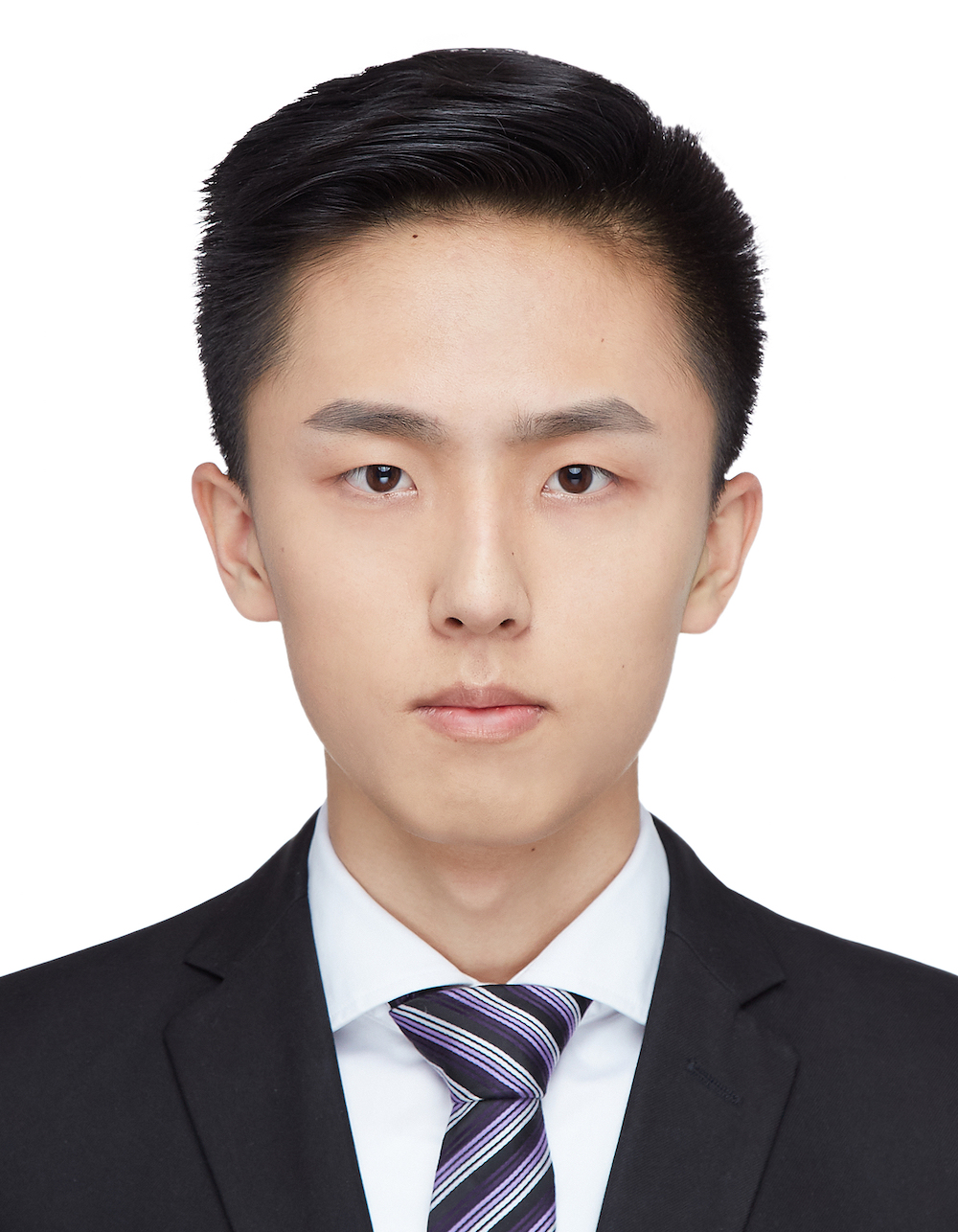}}]{Minghao Ning}
  received the B.S. degree in Vehicle Engineering from the Beijing Institute of Technology, Beijing, China, in 2020. He is currently pursuing a Ph.D. degree with the Department of Mechanical and Mechatronics Engineering, University of Waterloo. His research interests include autonomous driving, LiDAR perception, planning and control.
\end{IEEEbiography}
% \vspace{-0.4cm}

\begin{IEEEbiography}[{\includegraphics[width=1in,height=1.25in,clip,keepaspectratio]{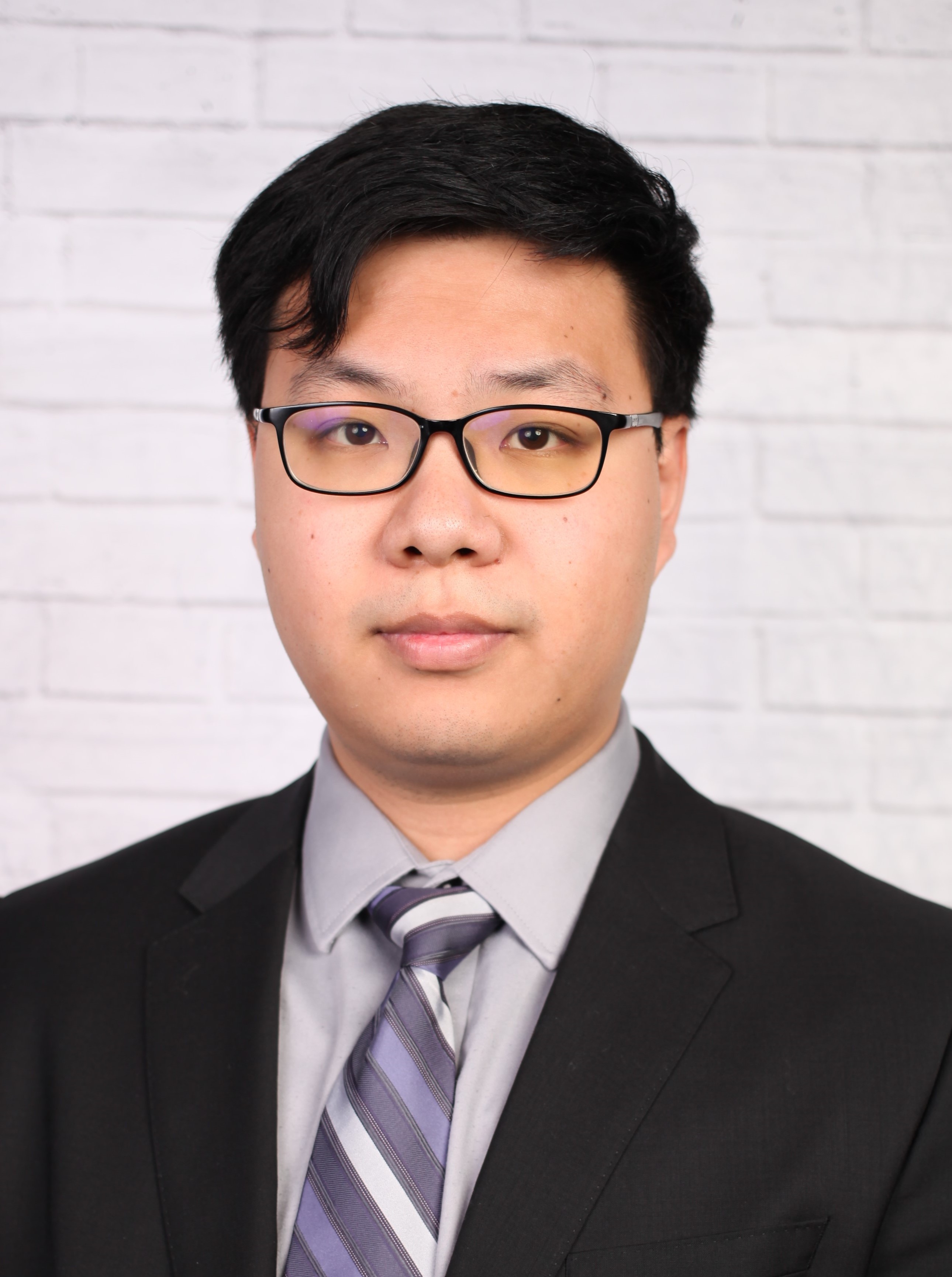}}]{Yufeng Yang}
received the B.Sc. degree in Mechanical Engineering with a minor in
Mechatronics from the University of Calgary in 2021. He is currently pursuing the Ph.D. degree with the Department of Mechanical and Mechatronics Engineering, University of Waterloo. His primary research interests include omnidirectional mobile robots, motion planning and control, and human–robot interaction.
\end{IEEEbiography}
% \vspace{-0.4cm}

\begin{IEEEbiography}[{\includegraphics[width=1in,height=1.25in,clip,keepaspectratio]{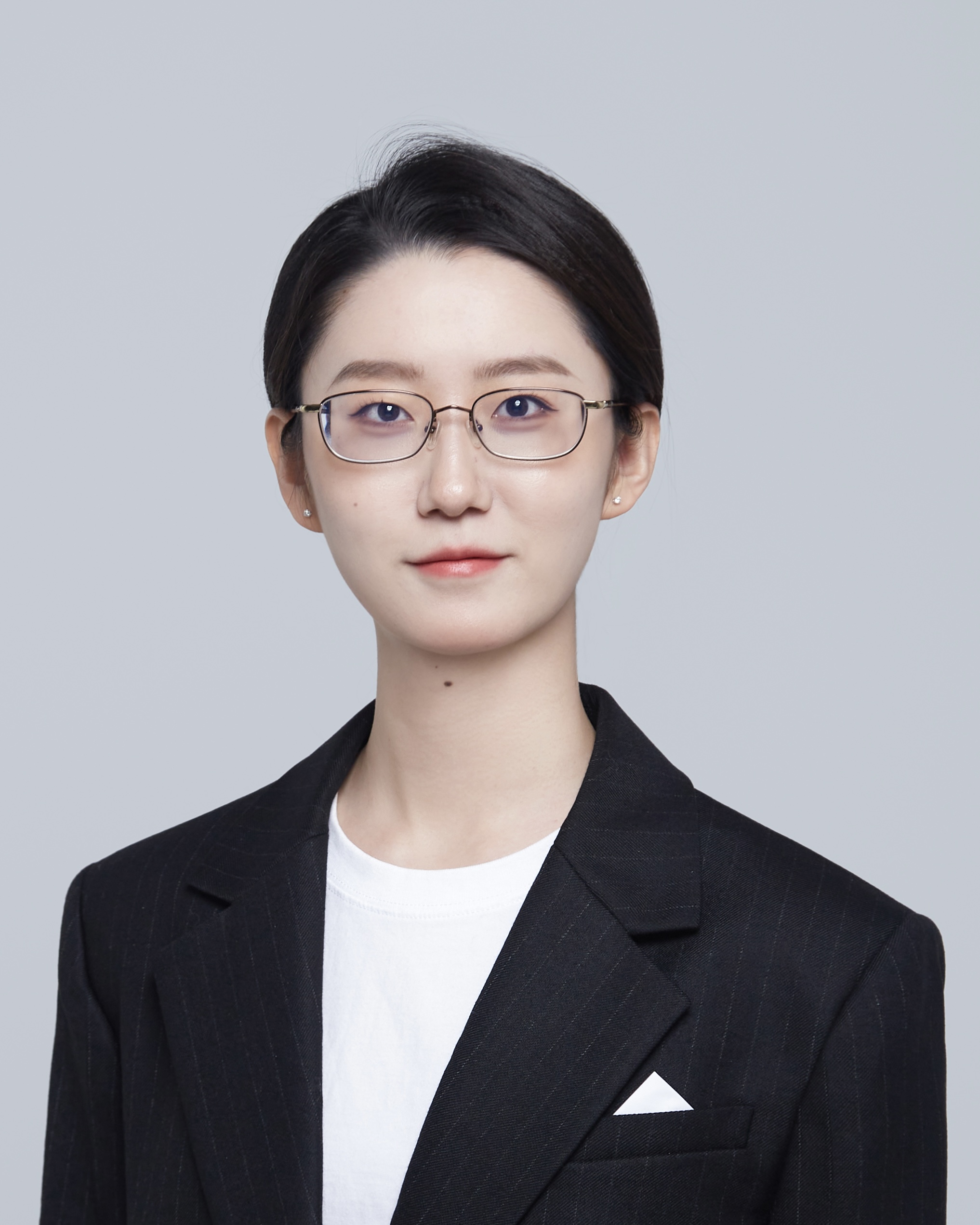}}]{Yukun Lu} is currently an Assistant Professor with the Department of Mechanical Engineering at the University of New Brunswick in Canada. She is also the Director of the Intelligent Mobility and Robotics Lab (IMRL). She earned her Ph.D. in Mechanical and Mechatronics Engineering from the University of Waterloo in 2023, where she continued as a Postdoctoral Researcher at the Mechatronic Vehicle Systems Lab (MVSL). She holds a B.Eng. in Vehicle Engineering with a minor in Business Administration, completed in 2018. Her background and future research interests include ground vehicle corner modules, intelligent robotic mobility, data-driven learning-based control strategies, vehicle dynamics and control, etc.
\end{IEEEbiography}
% \vspace{-0.4cm}

\begin{IEEEbiography}[{\includegraphics[width=1in,height=1.25in,clip,keepaspectratio]{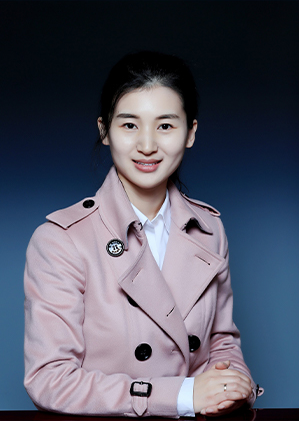}}]{Hong Wang}  (Senior Member, IEEE) received the
Ph.D. degree from the Beijing Institute of Technology, China, in 2015. From 2015 to 2019, she was a
Research Associate of mechanical and mechatronics
engineering with the University of Waterloo. She
is currently a Research Associate Professor with
Tsinghua University. She has published more than
60 articles on top international journals. Her research
interests include the safety of the on-board AI
algorithm, the safe decision-making for intelligent
vehicles, and the test and evaluation of SOTIF.
\end{IEEEbiography}
% \vspace{-0.4cm}

\begin{IEEEbiography}[{\includegraphics[width=1in,height=1.25in,clip,keepaspectratio]{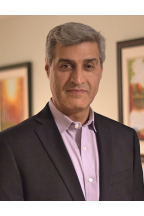}}]{Amir Khajepour} 
 (Senior Member, IEEE) is a professor of Mechanical and Mechatronics Engineering and the Director of the Mechatronic Vehicle Systems (MVS) Lab at the University of Waterloo. He held the Tier 1 Canada Research Chair in Mechatronic Vehicle Systems from 2008 to 2022 and the Senior NSERC/General Motors Industrial Research Chair in Holistic Vehicle Control from 2017 to 2022. His work has led to the training of over 150 PhD and MASc students, filing over 30 patents, publication of 600 research papers, numerous technology transfers, and the establishment of several start-up companies. He has been recognized with the Engineering Medal from Professional Engineering Ontario and is a fellow of the Engineering Institute of Canada, the American Society of Mechanical Engineering, and the Canadian Society of Mechanical Engineering.
\end{IEEEbiography}

\end{document}